\documentclass{article}

    \PassOptionsToPackage{numbers, compress}{natbib}

\usepackage[final]{neurips_2023}

\usepackage[utf8]{inputenc} 
\usepackage[T1]{fontenc}    
\usepackage[hidelinks]{hyperref}       
\usepackage{url}            
\usepackage{booktabs}       
\usepackage{amsfonts}       
\usepackage{nicefrac}       
\usepackage{microtype}      
\usepackage{xcolor}         

\usepackage{bm}
\usepackage{amsmath}
\usepackage{amssymb}
\usepackage{multirow}
\usepackage{subfigure}
\usepackage{array}
\usepackage{amsthm}
\usepackage{rotating}
\usepackage{pifont}
\usepackage{wrapfig}
\usepackage{caption} 

\newtheorem{definition}{Definition}
\newcommand{\figref}[1]{Figure \ref{#1}}
\newcommand{\tabref}[1]{Table \ref{#1}}
\newcommand{\secref}[1]{Section \ref{#1}}

\newcommand{\appendixref}[1]{Appendix}

\title{Towards Better Dynamic Graph Learning: \\New Architecture and Unified Library}

\author{%
  Le Yu, Leilei Sun\thanks{Corresponding Author.}, Bowen Du, Weifeng Lv\\
  State Key Laboratory of Software Development Environment\\
  School of Computer Science and Engineering\\
  Beihang University\\
  \texttt{\{yule,leileisun,dubowen,lwf\}@buaa.edu.cn} \\
}

\begin{document}

\maketitle

\begin{abstract}
We propose DyGFormer, a new Transformer-based architecture for dynamic graph learning. DyGFormer is conceptually simple and only needs to learn from nodes' historical first-hop interactions by: (\romannumeral1) a neighbor co-occurrence encoding scheme that explores the correlations of the source node and destination node based on their historical sequences; (\romannumeral2) a patching technique that divides each sequence into multiple patches and feeds them to Transformer, allowing the model to effectively and efficiently benefit from longer histories. We also introduce DyGLib, a unified library with standard training pipelines, extensible coding interfaces, and comprehensive evaluating protocols to promote reproducible, scalable, and credible dynamic graph learning research. By performing exhaustive experiments on thirteen datasets for dynamic link prediction and dynamic node classification tasks, we find that DyGFormer achieves state-of-the-art performance on most of the datasets, demonstrating its effectiveness in capturing nodes' correlations and long-term temporal dependencies. Moreover, some results of baselines are inconsistent with previous reports, which may be caused by their diverse but less rigorous implementations, showing the importance of DyGLib. All the used resources are publicly available at \url{https://github.com/yule-BUAA/DyGLib}.
\end{abstract}

\section{Introduction}
\label{section-1}

Dynamic graphs denote entities as nodes and represent their interactions as links with timestamps \cite{DBLP:journals/jmlr/KazemiGJKSFP20}, which can model many real-world scenarios such as social networks \cite{DBLP:conf/kdd/KumarZL19,DBLP:conf/wsdm/Song0WCZT19,alvarez2021evolutionary}, user-item interaction systems \cite{DBLP:conf/icdm/LiZWLWY20,DBLP:conf/cikm/FanLZX0Y21,DBLP:conf/www/YuWS0L22,zhang2022dynamic,yu2023continuous},
traffic networks \cite{DBLP:conf/ijcai/YuYZ18,DBLP:conf/ijcai/WuPLJZ19,DBLP:conf/aaai/GuoLFSW19,DBLP:conf/nips/0001YL0020,DBLP:journals/ijon/YuDHSHL21}, and physical systems \cite{DBLP:conf/nips/HuangS020,DBLP:conf/icml/Sanchez-Gonzalez20,DBLP:conf/iclr/PfaffFSB21}. In recent years, representation learning on dynamic graphs has become a trending research topic \cite{DBLP:journals/jmlr/KazemiGJKSFP20,skarding2021foundations,xue2022dynamic}. There are two main categories of existing methods: discrete-time \cite{DBLP:conf/aaai/ParejaDCMSKKSL20,DBLP:journals/kbs/GoyalCC20,DBLP:conf/wsdm/SankarWGZY20,DBLP:journals/corr/abs-2111-10447,DBLP:conf/kdd/YouDL22} and continuous-time \cite{DBLP:conf/iclr/XuRKKA20,DBLP:journals/corr/abs-2006-10637,DBLP:conf/iclr/WangCLL021,DBLP:conf/nips/SouzaMKG22,cong2023do}. In this paper, we focus on the latter approaches because they are more flexible and effective than the formers and are being increasingly investigated.

Despite the rapid development of dynamic graph learning methods, they still suffer from two limitations. Firstly, most of them independently compute the temporal representations of nodes in an interaction without exploiting nodes' correlations, which are often indicative of future interactions. Moreover, existing methods learn at the interaction level and thus only work for nodes with fewer interactions. When nodes have longer histories, they require sampling strategies to truncate the interactions for feasible calculations of the computationally expensive modules like graph convolutions \cite{DBLP:conf/iclr/TrivediFBZ19,DBLP:conf/iclr/XuRKKA20,DBLP:journals/corr/abs-2006-10637,DBLP:conf/sigir/0001GRTY20,DBLP:conf/cikm/ChangLW0FS020,DBLP:conf/sigmod/WangLLXYWWCYSG21}, temporal random walks \cite{DBLP:conf/iclr/WangCLL021,jin2022neural} and sequential models \cite{DBLP:journals/corr/abs-2105-07944,cong2023do}. Though some approaches use memory networks \cite{DBLP:journals/corr/WestonCB14,DBLP:conf/nips/SukhbaatarSWF15} to sequentially process interactions with affordable computational costs \cite{DBLP:conf/kdd/KumarZL19,DBLP:conf/iclr/TrivediFBZ19,DBLP:journals/corr/abs-2006-10637,DBLP:conf/sigir/0001GRTY20,DBLP:conf/sigmod/WangLLXYWWCYSG21,luo2022neighborhoodaware}, they are faced with the vanishing/exploding gradients due to the usage of recurrent neural networks \cite{DBLP:conf/icml/PascanuMB13,DBLP:journals/corr/abs-2105-07944}. Therefore, we conclude that \textit{previous methods lack the ability to capture either nodes' correlations or long-term temporal dependencies}. 

Secondly, the training pipelines of different methods are inconsistent and often lead to poor reproducibility. Also, existing methods are implemented by diverse frameworks (e.g., Pytorch \cite{DBLP:conf/nips/PaszkeGMLBCKLGA19}, Tensorflow \cite{DBLP:conf/osdi/AbadiBCCDDDGIIK16}, DGL \cite{DBLP:journals/corr/abs-1909-01315}, PyG \cite{DBLP:journals/corr/abs-1903-02428}, C++), making it time-consuming and difficult for researchers to quickly understand the algorithms and further dive into the core of dynamic graph learning. Although there exist some libraries for dynamic graph learning \cite{DBLP:journals/corr/abs-1811-10734,DBLP:conf/cikm/RozemberczkiSHP21,zhou2022tgl}, they mainly focus on dynamic network embedding methods \cite{DBLP:journals/corr/abs-1811-10734}, discrete-time graph learning methods \cite{DBLP:conf/cikm/RozemberczkiSHP21}, or engineering techniques for training on large-scale dynamic graphs \cite{zhou2022tgl} (elaborated in \secref{section-2}). Currently, we find that \textit{there are still no standard tools for continuous-time dynamic graph learning}.

In this paper, we aim to address the above drawbacks with two key technical contributions.

\textbf{We propose a new Transformer-based dynamic graph learning architecture (DyGFormer)}. DyGFormer is conceptually simple by solely learning from the sequences of nodes’ historical first-hop interactions. To be specific, DyGFormer is designed with a neighbor co-occurrence encoding scheme, which encodes the appearing frequencies of each neighbor in the sequences of the source and destination nodes to explicitly explore their correlations. In order to capture long-term temporal dependencies, DyGFormer splits each node's sequence into multiple patches and feeds them to Transformer \cite{DBLP:conf/nips/VaswaniSPUJGKP17}. This patching technique not only makes the model effectively benefit from longer histories via preserving local temporal proximities, but also efficiently reduces the computational complexity to a constant level that is irrelevant to the input sequence length.


\textbf{We present a unified continuous-time dynamic graph learning library (DyGLib)}. DyGLib is an open-source toolkit with standard training pipelines, extensible coding interfaces, and comprehensive evaluating strategies, aiming to foster standard, scalable, and reproducible dynamic graph learning research. DyGLib has integrated a variety of continuous-time dynamic graph learning methods as well as benchmark datasets from various domains. It trains all the methods via the same pipeline to eliminate the influence of different implementations and adopts a modularized design for developers to conveniently incorporate new datasets and algorithms based on their specific requirements. Moreover, DyGLib supports both dynamic link prediction and dynamic node classification tasks with exhaustive evaluating strategies to provide comprehensive comparisons of existing methods. 

To evaluate the model performance, we conduct extensive experiments based on DyGLib, including dynamic link prediction under transductive and inductive settings with three negative sampling strategies as well as dynamic node classification. From the results, we observe that: (\romannumeral1) DyGFormer outperforms existing methods on most datasets, demonstrating its superiority in capturing nodes' correlations and long-term temporal dependencies; (\romannumeral2) some findings of baselines are not in line with previous reports because of their varied pipelines and problematic implementations, which illustrates the necessity of introducing DyGLib.
We also provide an in-depth analysis of the neighbor co-occurrence encoding and patching technique for a better understanding of DyGFormer.

\section{Related Work}
\label{section-2}

\textbf{Dynamic Graph Learning}.
Representation learning on dynamic graphs has been widely studied in recent years \cite{DBLP:journals/jmlr/KazemiGJKSFP20,skarding2021foundations,xue2022dynamic}. 
Discrete-time methods manually divide the dynamic graph into a sequence of snapshots and apply static graph learning methods on each snapshot, which ignore the temporal order of nodes in each snapshot \cite{DBLP:conf/aaai/ParejaDCMSKKSL20,DBLP:journals/kbs/GoyalCC20,DBLP:conf/wsdm/SankarWGZY20,DBLP:journals/corr/abs-2111-10447,DBLP:conf/kdd/YouDL22}. In contrast, continuous-time methods directly learn on the whole dynamic graph with temporal graph neural networks \cite{DBLP:conf/iclr/TrivediFBZ19,DBLP:conf/iclr/XuRKKA20,DBLP:journals/corr/abs-2006-10637,DBLP:conf/sigir/0001GRTY20,DBLP:conf/cikm/ChangLW0FS020,DBLP:conf/sigmod/WangLLXYWWCYSG21}, memory networks \cite{DBLP:conf/kdd/KumarZL19,DBLP:conf/iclr/TrivediFBZ19,DBLP:journals/corr/abs-2006-10637,DBLP:conf/sigir/0001GRTY20,DBLP:conf/sigmod/WangLLXYWWCYSG21,luo2022neighborhoodaware}, temporal random walks \cite{DBLP:conf/iclr/WangCLL021,jin2022neural} or sequential models \cite{DBLP:journals/corr/abs-2105-07944,cong2023do}. Although insightful, most existing dynamic graph learning methods neglect the correlations between two nodes in an interaction. They also fail to handle nodes with longer interactions due to unaffordable computational costs of complex modules or issues in optimizing models (e.g., the vanishing/exploding gradients). In this paper, we propose a new Transformer-based architecture to show the necessity of capturing nodes' correlations and long-term temporal dependencies, which is achieved by two designs: a neighbor co-occurrence encoding scheme and a patching technique.

\textbf{Transformer-based Applications in Various Fields}. 
Transformer \cite{DBLP:conf/nips/VaswaniSPUJGKP17} is an innovative model that employs the self-attention mechanism to handle sequential data, which has been successfully applied in a variety of domains, such as natural language processing \cite{DBLP:conf/naacl/DevlinCLT19,DBLP:journals/corr/abs-1907-11692,DBLP:conf/nips/BrownMRSKDNSSAA20}, computer vision \cite{DBLP:conf/eccv/CarionMSUKZ20,DBLP:conf/iclr/DosovitskiyB0WZ21,DBLP:conf/iccv/LiuL00W0LG21} and time series forecasting \cite{DBLP:conf/nips/LiJXZCWY19,DBLP:conf/aaai/ZhouZPZLXZ21,DBLP:conf/nips/WuXWL21}. The idea of dividing the original data into patches as inputs of the Transformer has been attempted in some studies. ViT \cite{DBLP:conf/iclr/DosovitskiyB0WZ21} splits an image
into multiple patches and feeds the sequence of patches' linear embeddings into a Transformer, which achieves surprisingly good performance on image classification. PatchTST \cite{nie2023a} divides a time series into subseries-level patches and calculates the patches by a channel-independent Transformer for long-term multivariate time series forecasting. In this work, we propose a patching technique to learn on dynamic graphs, which can provide our approach with the ability to handle nodes with longer histories.

\textbf{Graph Learning Library}. Currently, there exist many libraries for static graphs \cite{DBLP:journals/corr/abs-1806-01261,DBLP:journals/corr/abs-1909-01315,DBLP:journals/corr/abs-1903-02428,DBLP:conf/nips/HuFZDRLCL20,DBLP:journals/corr/abs-2103-00959,DBLP:conf/icse/LiXCZL21,DBLP:journals/jmlr/LiuLWXYGYXZLYLF21}, but few for dynamic graph learning \cite{DBLP:journals/corr/abs-1811-10734,DBLP:conf/cikm/RozemberczkiSHP21,zhou2022tgl}. 
DynamicGEM \cite{DBLP:journals/corr/abs-1811-10734} focuses on dynamic graph embedding methods, which just consider the graph topology and cannot leverage node features. PyTorch Geometric Temporal \cite{DBLP:conf/cikm/RozemberczkiSHP21} implements discrete-time algorithms for spatiotemporal signal processing and is mainly applicable for nodes with aligned historical observations. TGL \cite{zhou2022tgl} trains on large-scale dynamic graphs with some engineering tricks. Though TGL has integrated some continuous-time methods, they are somewhat out-of-date, resulting in the lack of state-of-the-art models. Moreover, TGL is implemented by both PyTorch and C++, which needs additional compilation and increases the usage difficulty. In this paper, we present a unified continuous-time dynamic graph learning library with thorough baselines, diverse datasets, extensible implementations, and comprehensive evaluations to facilitate dynamic graph learning research.

\section{Preliminaries}
\label{section-3}

\begin{definition}
    \textbf{Dynamic Graph}. We represent a dynamic graph as a sequence of non-decreasing chronological interactions $\mathcal{G}=\left\{\left(u_1,v_1,t_1\right), \left(u_2,v_2,t_2\right), \cdots \right\}$ with $0 \leq t_1 \leq t_2 \leq \cdots$, where $u_i, v_i \in \mathcal{N}$ denote the source node and destination node of the $i$-th link at timestamp $t_i$.
    $\mathcal{N}$ is the set of all the nodes. Each node $u \in \mathcal{N}$ can be associated with node feature $\bm{x}_u \in \mathbb{R}^{d_N}$, and each interaction $\left(u,v,t\right)$ has link feature $\bm{e}_{u,v}^t \in \mathbb{R}^{d_E}$. $d_N$ and $d_E$ denote the dimensions of the node feature and link feature. If the graph is non-attributed, we simply set the node feature and link feature to zero vectors, i.e., $\bm{x}_u=\bm{0}$ and $\bm{e}_{u,v}^t=\bm{0}$. 
\end{definition}

\begin{definition}
    \textbf{Problem Formalization}. Given the source node $u$, destination node $v$, timestamp $t$, and historical interactions before $t$, i.e., $\left\{\left(u^\prime,v^\prime,t^\prime\right) | t^\prime < t \right\}$, representation learning on dynamic graph aims to design a model to learn time-aware representations $\bm{h}_u^t \in \mathbb{R}^d$ and $\bm{h}_v^t \in \mathbb{R}^d$ for $u$ and $v$ with $d$ as the dimension. We validate the effectiveness of the learned representations via two common tasks in dynamic graph learning: (\romannumeral1) dynamic link prediction, which predicts whether $u$ and $v$ are connected at $t$; (\romannumeral2) dynamic node classification, which infers the state of $u$ or $v$ at $t$.
\end{definition}

\section{New Architecture and Unified Library}
\label{section-4}

\subsection{DyGFormer: Transformer-based Architecture for Dynamic Graph Learning}
\begin{figure}[!ht]
    \centering
    \includegraphics[width=1.0\columnwidth]{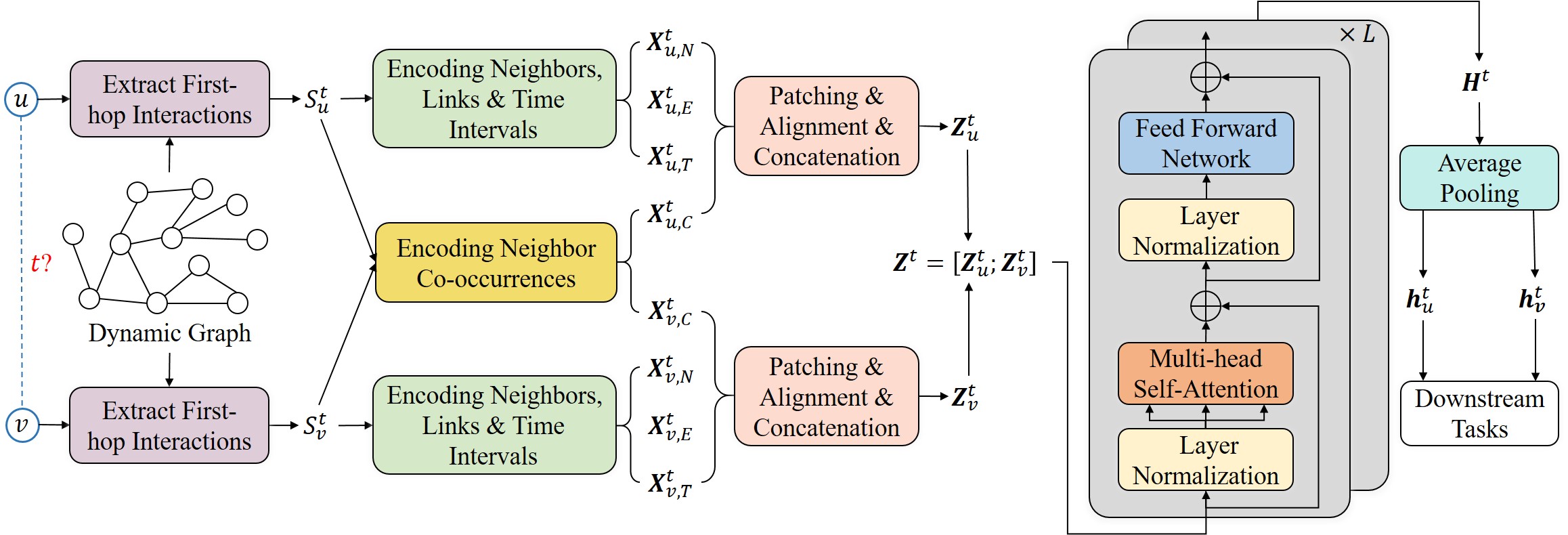}
    \caption{Framework of the proposed model.}
    \label{fig:DyGFormer_framework}
\end{figure}

The framework of our DyGFormer is shown in \figref{fig:DyGFormer_framework}, which employs Transformer \cite{DBLP:conf/nips/VaswaniSPUJGKP17} as the backbone. Given an interaction $\left(u,v,t\right)$, we first extract historical first-hop interactions of source node $u$ and destination node $v$ before timestamp $t$ and obtain two interaction sequences $\mathcal{S}_u^t$ and $\mathcal{S}_v^t$. Next, in addition to computing the encodings of neighbors, links, and time intervals for each sequence, we also encode the frequencies of every neighbor's appearances in both $\mathcal{S}_u^t$ and $\mathcal{S}_v^t$ to exploit the correlations between $u$ and $v$, resulting in four encoding sequences for $u$/$v$ in total. Then, we divide each encoding sequence into multiple patches and feed all the patches into a Transformer for capturing long-term temporal dependencies. Finally, the outputs of the Transformer are averaged to derive time-aware representations of $u$ and $v$ at timestamp $t$ (i.e., $\bm{h}_u^t$ and $\bm{h}_v^t$), which can be applied in various downstream tasks like dynamic link prediction and dynamic node classification.

\textbf{Learning from Historical First-hop Interactions}. Unlike most previous methods that require nodes' historical interactions from multiple hops (e.g., DyRep \cite{DBLP:conf/iclr/TrivediFBZ19}, TGAT \cite{DBLP:conf/iclr/XuRKKA20}, TGN \cite{DBLP:journals/corr/abs-2006-10637}, CAWN \cite{DBLP:conf/iclr/WangCLL021}), we only learn from the sequences of nodes' historical first-hop interactions, turning the dynamic graph learning task into a simpler sequence learning problem. Mathematically, given an interaction $\left(u,v,t\right)$, for source node $u$ and destination node $v$, we obtain the sequences that involve first-hop interactions of $u$ and $v$ before timestamp $t$, which are denoted by $\mathcal{S}_u^t=\left\{\left(u,u^\prime,t^\prime\right) | t^\prime < t \right\} \cup \left\{\left(u^\prime,u,t^\prime\right) | t^\prime < t \right\}$ and $\mathcal{S}_v^t=\left\{\left(v,v^\prime,t^\prime\right) | t^\prime < t \right\} \cup \left\{\left(v^\prime,v,t^\prime\right) | t^\prime < t \right\}$, respectively.

\textbf{Encoding Neighbors, Links, and Time Intervals}. For source node $u$, we retrieve the features of involved neighbors and links in sequence $\mathcal{S}_u^t$ based on the given features to represent their encodings, which are denoted by $\bm{X}_{u,N}^t \in \mathbb{R}^{|\mathcal{S}_u^t| \times d_N}$ and $\bm{X}_{u,E}^t \in \mathbb{R}^{|\mathcal{S}_u^t| \times d_E}$. Following \cite{DBLP:conf/iclr/XuRKKA20}, we learn the periodic temporal patterns by encoding the time interval $\Delta t^\prime = t - t^\prime$ via $\sqrt{\frac{1}{d_T}}\left[\cos\left(w_1 \Delta t^\prime \right), \sin\left(w_1 \Delta t^\prime \right), \cdots, \cos\left(w_{d_T} \Delta t^\prime \right), \sin \left(w_{d_T} \Delta t^\prime \right)\right]$, where $w_1,\cdots,w_{d_T}$ are trainable parameters. $d_T$ is the encoding dimension. The time interval encodings of interactions in $\mathcal{S}_u^t$ is denoted by $\bm{X}_{u,T}^t \in \mathbb{R}^{|\mathcal{S}_u^t| \times d_T}$. We use the same process to get the corresponding encodings for destination node $v$, i.e., $\bm{X}_{v,N}^t \in \mathbb{R}^{|\mathcal{S}_v^t| \times d_N}$, $\bm{X}_{v,E}^t \in \mathbb{R}^{|\mathcal{S}_v^t| \times d_E}$, and $\bm{X}_{v,T}^t \in \mathbb{R}^{|\mathcal{S}_v^t| \times d_T}$.

\textbf{Neighbor Co-occurrence Encoding Scheme}. Existing methods separately compute representations of node $u$ and $v$ without modeling their correlations. We present a neighbor co-occurrence encoding scheme to tackle this issue, which assumes the appearing frequency of a neighbor in a sequence indicates its importance, and the occurrences of a neighbor in sequences of $u$ and $v$ (i.e., co-occurrence) could reflect the correlations between $u$ and $v$. That is to say, if $u$ and $v$ have more common historical neighbors in their sequences, they are more likely to interact in the future. 

Formally, for each neighbor in the interaction sequence $\mathcal{S}_u^t$ and $\mathcal{S}_v^t$, we count its occurrences in both $\mathcal{S}_u^t$ and $\mathcal{S}_v^t$, and derive a two-dimensional vector. By packing the vectors of all the neighbors together, we can get the neighbor co-occurrence features for $u$ and $v$, which are represented by $\bm{C}_{u}^t \in \mathbb{R}^{|\mathcal{S}_u^t| \times 2}$ and $\bm{C}_{v}^t \in \mathbb{R}^{|\mathcal{S}_v^t| \times 2}$. For example, suppose the historical neighbors of $u$ and $v$ are $\left\{a, b, a\right\}$ and $\left\{b, b, a, c\right\}$. The appearing frequencies of $a$, $b$, and $c$ in $u$/$v$'s historical interactions are 2/1, 1/2, and 0/1, respectively. Therefore, the neighbor co-occurrence features of $u$ and $v$ are denoted by $\bm{C}_u^t=\left[\left[2, 1\right], \left[1, 2\right],\left[2, 1\right]\right]^\top$ and $\bm{C}_v^t=\left[\left[1, 2\right], \left[1, 2\right],\left[2, 1\right],\left[0, 1\right]\right]^\top$. Then, we apply a function $f\left(\cdot\right)$ to encode the neighbor co-occurrence features by 
\begin{equation}
\label{equ:node_co_occurrence_encoding}
\begin{split}
   \bm{X}_{*,C}^t = f\left(\bm{C}_*^t\left[:,0\right]\right) + f\left(\bm{C}_*^t\left[:,1\right]\right) \in \mathbb{R}^{|\mathcal{S}_*^t| \times d_C},
\end{split}
\end{equation}
where $*$ could be $u$ or $v$. The input and output dimensions of $f\left(\cdot\right)$ are 1 and $d_C$. In this paper, we implement $f\left(\cdot\right)$ by a two-layer perceptron with ReLU activation \cite{DBLP:conf/icml/NairH10}.
It is important to note that the neighbor co-occurrence encoding scheme is general and can be easily integrated into some dynamic graph learning methods for better results. We will demonstrate its generalizability in \secref{section-5-node_cooccurrence_generalizability}. 

\textbf{Patching Technique}. Instead of focusing on the interaction level, we divide the encoding sequence into multiple non-overlapping patches to break through the bottleneck of existing methods in capturing long-term temporal dependencies. Let $P$ denote the patch size. Each patch is composed of $P$ temporally adjacent interactions with flattened encodings and can preserve local temporal proximities. Take the patching of $\bm{X}_{u,N}^t \in \mathbb{R}^{|\mathcal{S}_u^t| \times d_N}$ as an example. $\bm{X}_{u,N}^t$ will be divided into $l_u^t =\lceil \frac{|\mathcal{S}_u^t|}{P} \rceil$ patches in total (note that we will pad $\bm{X}_{u,N}^t$ if its length $|\mathcal{S}_u^t|$ cannot be divided by $P$), and the patched encoding is represented by $\bm{M}_{u,N}^t \in \mathbb{R}^{l_u^t \times d_N \cdot P}$. Similarly, we can also get the patched encodings $\bm{M}_{u,E}^t \in \mathbb{R}^{l_u^t  \times d_E \cdot P}$, $\bm{M}_{u,T}^t \in \mathbb{R}^{l_u^t \times d_T \cdot P}$, $\bm{M}_{u,C}^t \in \mathbb{R}^{l_u^t  \times d_C \cdot P}$, 
$\bm{M}_{v,N}^t \in \mathbb{R}^{l_v^t  \times d_N \cdot P}$, $\bm{M}_{v,E}^t \in \mathbb{R}^{l_v^t  \times d_E \cdot P}$, $\bm{M}_{v,T}^t \in \mathbb{R}^{l_v^t  \times d_T \cdot P}$, and $\bm{M}_{v,C}^t \in \mathbb{R}^{l_v^t  \times d_C \cdot P}$. Note that when $|\mathcal{S}_u^t|$ becomes longer, we will correspondingly increase $P$, making the number of patches (i.e., $l_u^t$ and $l_v^t$) at a constant level to reduce the computational cost.

\textbf{Transformer Encoder}. We first align the patched encodings to the same dimension $d$ with trainable weight $\bm{W}_* \in \mathbb{R}^{d_* \cdot P \times d}$ and $\bm{b}_* \in \mathbb{R}^{d}$ to obtain $\bm{Z}_{u,*}^t \in \mathbb{R}^{l_u^t \times d}$ and $\bm{Z}_{v,*}^t \in \mathbb{R}^{l_v^t \times d}$, where $*$ could be $N$, $E$, $T$ or $C$. To be specific, the alignments are realized by
\begin{equation}
    \label{equ:projection_layer}
    \bm{Z}_{u,*}^t = \bm{M}_{u,*}^t \bm{W}_* + \bm{b}_* \in \mathbb{R}^{l_u^t \times d}, \bm{Z}_{v,*}^t = \bm{M}_{v,*}^t \bm{W}_* + \bm{b}_* \in \mathbb{R}^{l_v^t \times d}.
\end{equation}
Then, we concatenate the aligned encodings of $u$ and $v$, and get $\bm{Z}_{u}^t = \bm{Z}_{u,N}^t \| \bm{Z}_{u,E}^t \| \bm{Z}_{u,T}^t \| \bm{Z}_{u,C}^t \in \mathbb{R}^{l_u^t \times 4d}$ and $\bm{Z}_{v}^t = \bm{Z}_{v,N}^t \| \bm{Z}_{v,E}^t \| \bm{Z}_{v,T}^t \| \bm{Z}_{v,C}^t \in \mathbb{R}^{l_v^t \times 4d}$. 

Next, we employ a Transformer encoder to capture the temporal dependencies, which is built by stacking $L$ Multi-head Self-Attention (MSA) and Feed-Forward Network (FFN) blocks. The residual connection \cite{DBLP:conf/cvpr/HeZRS16} is employed after every block. We follow \cite{DBLP:conf/iclr/DosovitskiyB0WZ21} by using GELU \cite{hendrycks2016gaussian} instead of ReLU \cite{DBLP:conf/icml/NairH10} between the two-layer perception in each FFN block and applying Layer Normalization (LN) \cite{ba2016layer} before each block rather than after.
Instead of individually processing $\bm{Z}_{u}^t$ and $\bm{Z}_{v}^t$, our Transformer encoder takes the stacked $\bm{Z}^t=\left[\bm{Z}_{u}^t; \bm{Z}_{v}^t\right] \in \mathbb{R}^{(l_u^t + l_v^t) \times 4d}$ as inputs, aiming to learn the temporal dependencies within and across the sequences of $u$ and $v$. The calculation process is
\begin{gather} 
\label{equ:transformer}
   \text{Attention}\left(\bm{Q}, \bm{K}, \bm{V}\right)= \text{Softmax}\left(\frac{\bm{Q} \bm{K}^\top}{\sqrt{d_k}}\right) \bm{V},\\
   \text{FFN}\left(\bm{O}, \bm{W}_1, \bm{b}_1, \bm{W}_2, \bm{b}_2\right)= \text{GELU}\left(\bm{O}\bm{W}_1+\bm{b}_1\right)\bm{W}_2 + \bm{b}_2,\\
   \bm{O}_i^{t,l} = \text{Attention}\left( \text{LN}(\bm{Z}^{t,l-1}) \bm{W}_{Q,i}^l, \text{LN}(\bm{Z}^{t,l-1}) \bm{W}_{K,i}^l, \text{LN}(\bm{Z}^{t,l-1}) \bm{W}_{V,i}^l \right),\\
   \bm{O}^{t,l} = \text{MSA}\left(\bm{Z}^{t,l-1}\right) + \bm{Z}^{t,l-1} = \left(\bm{O}_1^{t,l} \| \cdots \| \bm{O}_I^{t,l}\right)\bm{W}_O^l + \bm{Z}^{t,l-1},\\
   \bm{Z}^{t,l} = \text{FFN}\left(\text{LN}\left(\bm{O}^{t,l}\right), \bm{W}_1^l, \bm{b}_1^l, \bm{W}_2^l, \bm{b}_2^l\right) + \bm{O}^{t,l}.
\end{gather}
$\bm{W}_{Q,i}^l \in \mathbb{R}^{4d \times d_k}$, $\bm{W}_{K,i}^l \in \mathbb{R}^{4d \times d_k}$, $\bm{W}_{V,i}^l \in \mathbb{R}^{4d \times d_v}$, $\bm{W}_O^l \in \mathbb{R}^{I \cdot d_v \times 4d}$, $\bm{W}_1^l \in \mathbb{R}^{4d \times 16d}$, $\bm{b}_1^l \in \mathbb{R}^{16d}$, $\bm{W}_2^l \in \mathbb{R}^{16d \times 4d}$ and $\bm{b}_2^l \in \mathbb{R}^{4d}$ are trainable parameters at the $l$-th layer. We set $d_k=d_v=4d/I$ with $I$ as the number of attention heads. The input of the first layer is $\bm{Z}^{t,0} = \bm{Z}^t \in \mathbb{R}^{(l_u^t + l_v^t) \times 4d}$, and the output of the $L$-th layer is denoted by $\bm{H}^t=\bm{Z}^{t,L} \in \mathbb{R}^{(l_u^t + l_v^t) \times 4d}$.

\textbf{Time-aware Node Representation}. The time-aware representations of node $u$ and $v$ at timestamp $t$ are derived by averaging their related representations in $\bm{H}^t$ with an output layer,
\begin{equation}
\label{equ:final_temporal_representation}
\begin{split}
   \bm{h}_u^{t} & = \text{MEAN}\left(\bm{H}^t[:l_u^t,:]\right) \bm{W}_{out} + \bm{b}_{out} \in \mathbb{R}^{d_{out}},\\
   \bm{h}_v^{t} & = \text{MEAN}\left(\bm{H}^t[l_u^t:l_u^t + l_v^t,:]\right) \bm{W}_{out} + \bm{b}_{out} \in \mathbb{R}^{d_{out}},\\
\end{split}
\end{equation}
where $\bm{W}_{out} \in \mathbb{R}^{4d \times d_{out}}$ and $\bm{b}_{out} \in \mathbb{R}^{d_{out}}$ are trainable weights with $d_{out}$ as the output dimension.

\subsection{DyGLib: Unified Library for Continuous-Time Dynamic Graph Learning}
We introduce a unified library with standard training pipelines, extensible coding interfaces, and comprehensive evaluating strategies for reproducible, scalable, and credible continuous-time dynamic graph learning research. The overall procedure of DyGLib is shown in \figref{fig:procedure_DyGLib} in \secref{section-appendix-DyGLib-procedure}.

\textbf{Standard Training Pipelines}. To eliminate the influence of different training pipelines in previous studies, we unify the data format, create a customized data loader, and train all the methods with the same model trainers. Our standard training pipelines guarantee reproducible performance and enable users to quickly identify the key components of different models. Researchers only need to focus on designing the model architecture without considering other irrelevant implementation details.

\textbf{Extensible Coding Interfaces}. We provide extensible coding interfaces for the datasets and algorithms, which are all implemented by PyTorch. These scalable designs enable users to incorporate new datasets and popular models based on their specific requirements, which can significantly reduce the usage difficulty for beginners and allow experts to conveniently validate new ideas. Currently, DyGLib has integrated thirteen datasets from various domains and nine continuous-time dynamic graph learning methods. It is worth noticing that we also found some issues in previous implementations and have fixed them in DyGLib (see details in \secref{section-appendix-issues-existing-methods}).

\textbf{Comprehensive Evaluating Protocols}. DyGLib supports both transductive/inductive dynamic link prediction and dynamic node classification tasks. Most previous works evaluate their methods on the dynamic link prediction task with the random negative sampling strategy but a few models already reach saturation performance under such a strategy, making it hard to distinguish more advanced designs. For more reliable comparisons, we adopt three strategies (i.e., random, historical, and inductive negative sampling strategies) in \cite{poursafaei2022towards} to comprehensively evaluate the model performance.

\section{Experiments}
\label{section-5}
In this section, we report the results of various approaches by using DyGLib. We show the superiority of DyGFormer over existing methods and also give an in-depth analysis of DyGFormer.

\subsection{Experimental Settings}
\textbf{Datasets and Baselines}. We experiment with thirteen datasets (Wikipedia, Reddit, MOOC, LastFM, Enron, Social Evo., UCI, Flights, Can. Parl., US Legis., UN Trade, UN Vote, and Contact), which are collected by \cite{poursafaei2022towards} and cover diverse domains. Details of the datasets are shown in \secref{section-appendix-descriptions_datasets}. We compare DyGFormer with eight popular continuous-time dynamic graph learning baselines that are based on graph convolutions, memory networks, random walks, and sequential models, including JODIE \cite{DBLP:conf/kdd/KumarZL19}, DyRep \cite{DBLP:conf/iclr/TrivediFBZ19}, TGAT \cite{DBLP:conf/iclr/XuRKKA20}, TGN \cite{DBLP:journals/corr/abs-2006-10637}, CAWN \cite{DBLP:conf/iclr/WangCLL021}, EdgeBank \cite{poursafaei2022towards}, TCL \cite{DBLP:journals/corr/abs-2105-07944}, and GraphMixer \cite{cong2023do}. We give the descriptions of baselines in \secref{section-appendix-descriptions_baselines}.

\textbf{Evaluation Tasks and Metrics}.
We follow \cite{DBLP:conf/iclr/XuRKKA20,DBLP:journals/corr/abs-2006-10637,DBLP:conf/iclr/WangCLL021,poursafaei2022towards} to evaluate models for dynamic link prediction, which predicts the probability of a link occurring between two given nodes at a specific time. This task has two settings: the transductive setting aims to predict future links between nodes that are observed during training, and the inductive setting predicts future links between unseen nodes. We use a multi-layer perceptron to take the concatenated representations of two nodes as inputs and return the probability of a link as the output. Average Precision (AP) and Area Under the Receiver Operating Characteristic Curve (AUC-ROC) are adopted as the evaluation metrics. We adopt random (rnd), historical (hist), and inductive (ind) negative sampling strategies in \cite{poursafaei2022towards} for evaluation, where the latter two strategies are more challenging. Please refer to \cite{poursafaei2022towards} for more details. We also follow \cite{DBLP:conf/iclr/XuRKKA20,DBLP:journals/corr/abs-2006-10637} to conduct dynamic node classification, which estimates the state of a node in a given interaction at a specific time. A multi-layer perceptron is employed to map the node representations to the labels. We use AUC-ROC as the evaluation metric due to the label imbalance. For both tasks, we chronologically split each dataset with the ratio of 70\%/15\%/15\% for training/validation/testing.

\textbf{Model Configurations}. For baselines, in addition to following their official settings, we also perform an exhaustive grid search to find the optimal configurations of some critical hyperparameters for more reliable comparisons. As DyGFormer can access longer histories, we vary each node's input sequence length from 32 to 4096 by a factor of 2. To keep the computational complexity at a constant level that is irrelevant to the input length, we correspondingly increase the patch size from 1 to 128. Please see \secref{section-appendix-configurations} for the detailed configurations of different models.

\textbf{Implementation Details}. For both tasks, we optimize all models (i.e., excluding EdgeBank which has no trainable parameters) by Adam \cite{DBLP:journals/corr/KingmaB14} and use supervised binary cross-entropy loss as the objective function. We train the models for 100 epochs and use the early stopping strategy with a patience of 20. We select the model that achieves the best performance on the validation set for testing. We set the learning rate and batch size to 0.0001 and 200 for all the methods on all the datasets. We run the methods five times with seeds from 0 to 4 and report the average performance to eliminate deviations. Experiments are conducted on an Ubuntu machine equipped with one Intel(R) Xeon(R) Gold 6130 CPU @ 2.10GHz with 16 physical cores. The GPU device is NVIDIA Tesla T4 with 15 GB memory. 

\subsection{Performance Comparisons and Discussions}
\begin{table}[!htbp]
\centering
\caption{AP for transductive dynamic link prediction with random, historical, and inductive negative sampling strategies. NSS is the abbreviation of Negative Sampling Strategies.}
\label{tab:average_precision_transductive_dynamic_link_prediction}
\resizebox{1.01\textwidth}{!}
{
\setlength{\tabcolsep}{0.9mm}
{
\begin{tabular}{c|c|ccccccccc}
\hline
NSS                    & Datasets    & JODIE        & DyRep        & TGAT         & TGN          & CAWN         & EdgeBank     & TCL          & GraphMixer   & DyGFormer    \\ \hline
\multirow{14}{*}{rnd}  & Wikipedia   & 96.50 $\pm$ 0.14 & 94.86 $\pm$ 0.06 & 96.94 $\pm$ 0.06 & 98.45 $\pm$ 0.06 & \underline{98.76 $\pm$ 0.03} & 90.37 $\pm$ 0.00 & 96.47 $\pm$ 0.16 & 97.25 $\pm$ 0.03 & \textbf{99.03 $\pm$ 0.02} \\
                       & Reddit      & 98.31 $\pm$ 0.14 & 98.22 $\pm$ 0.04 & 98.52 $\pm$ 0.02 & 98.63 $\pm$ 0.06 & \underline{99.11 $\pm$ 0.01} & 94.86 $\pm$ 0.00 & 97.53 $\pm$ 0.02 & 97.31 $\pm$ 0.01 & \textbf{99.22 $\pm$ 0.01} \\
                       & MOOC        & 80.23 $\pm$ 2.44 & 81.97 $\pm$ 0.49 & 85.84 $\pm$ 0.15 & \textbf{89.15 $\pm$ 1.60} & 80.15 $\pm$ 0.25 & 57.97 $\pm$ 0.00 & 82.38 $\pm$ 0.24 & 82.78 $\pm$ 0.15 & \underline{87.52 $\pm$ 0.49} \\
                       & LastFM      & 70.85 $\pm$ 2.13 & 71.92 $\pm$ 2.21 & 73.42 $\pm$ 0.21 & 77.07 $\pm$ 3.97 & \underline{86.99 $\pm$ 0.06} & 79.29 $\pm$ 0.00 & 67.27 $\pm$ 2.16 & 75.61 $\pm$ 0.24 & \textbf{93.00 $\pm$ 0.12} \\
                       & Enron       & 84.77 $\pm$ 0.30 & 82.38 $\pm$ 3.36 & 71.12 $\pm$ 0.97 & 86.53 $\pm$ 1.11 & \underline{89.56 $\pm$ 0.09} & 83.53 $\pm$ 0.00 & 79.70 $\pm$ 0.71 & 82.25 $\pm$ 0.16 & \textbf{92.47 $\pm$ 0.12} \\
                       & Social Evo. & 89.89 $\pm$ 0.55 & 88.87 $\pm$ 0.30 & 93.16 $\pm$ 0.17 & \underline{93.57 $\pm$ 0.17} & 84.96 $\pm$ 0.09 & 74.95 $\pm$ 0.00 & 93.13 $\pm$ 0.16 & 93.37 $\pm$ 0.07 & \textbf{94.73 $\pm$ 0.01} \\
                       & UCI         & 89.43 $\pm$ 1.09 & 65.14 $\pm$ 2.30 & 79.63 $\pm$ 0.70 & 92.34 $\pm$ 1.04 & \underline{95.18 $\pm$ 0.06} & 76.20 $\pm$ 0.00 & 89.57 $\pm$ 1.63 & 93.25 $\pm$ 0.57 & \textbf{95.79 $\pm$ 0.17} \\
                       & Flights     & 95.60 $\pm$ 1.73 & 95.29 $\pm$ 0.72 & 94.03 $\pm$ 0.18 & 97.95 $\pm$ 0.14 & \underline{98.51 $\pm$ 0.01} & 89.35 $\pm$ 0.00 & 91.23 $\pm$ 0.02 & 90.99 $\pm$ 0.05 & \textbf{98.91 $\pm$ 0.01} \\
                       & Can. Parl.  & 69.26 $\pm$ 0.31 & 66.54 $\pm$ 2.76 & 70.73 $\pm$ 0.72 & 70.88 $\pm$ 2.34 & 69.82 $\pm$ 2.34 & 64.55 $\pm$ 0.00 & 68.67 $\pm$ 2.67 & \underline{77.04 $\pm$ 0.46} & \textbf{97.36 $\pm$ 0.45} \\
                       & US Legis.   & 75.05 $\pm$ 1.52 & \underline{75.34 $\pm$ 0.39} & 68.52 $\pm$ 3.16 & \textbf{75.99 $\pm$ 0.58} & 70.58 $\pm$ 0.48 & 58.39 $\pm$ 0.00 & 69.59 $\pm$ 0.48 & 70.74 $\pm$ 1.02 & 71.11 $\pm$ 0.59 \\
                       & UN Trade    & 64.94 $\pm$ 0.31 & 63.21 $\pm$ 0.93 & 61.47 $\pm$ 0.18 & 65.03 $\pm$ 1.37 & \underline{65.39 $\pm$ 0.12} & 60.41 $\pm$ 0.00 & 62.21 $\pm$ 0.03 & 62.61 $\pm$ 0.27 & \textbf{66.46 $\pm$ 1.29} \\
                       & UN Vote     & \underline{63.91 $\pm$ 0.81} & 62.81 $\pm$ 0.80 & 52.21 $\pm$ 0.98 & \textbf{65.72 $\pm$ 2.17} & 52.84 $\pm$ 0.10 & 58.49 $\pm$ 0.00 & 51.90 $\pm$ 0.30 & 52.11 $\pm$ 0.16 & 55.55 $\pm$ 0.42 \\
                       & Contact     & 95.31 $\pm$ 1.33 & 95.98 $\pm$ 0.15 & 96.28 $\pm$ 0.09 & \underline{96.89 $\pm$ 0.56} & 90.26 $\pm$ 0.28 & 92.58 $\pm$ 0.00 & 92.44 $\pm$ 0.12 & 91.92 $\pm$ 0.03 & \textbf{98.29 $\pm$ 0.01} \\ \cline{2-11} 
                       & Avg. Rank   & 5.08         & 5.85         & 5.69         & \underline{2.54}         & 4.31         & 7.54         & 6.92         & 5.46         & \textbf{1.62}         \\ \hline
\multirow{14}{*}{hist} & Wikipedia   & 83.01 $\pm$ 0.66 & 79.93 $\pm$ 0.56 & 87.38 $\pm$ 0.22 & 86.86 $\pm$ 0.33 & 71.21 $\pm$ 1.67 & 73.35 $\pm$ 0.00 & \underline{89.05 $\pm$ 0.39} & \textbf{90.90 $\pm$ 0.10} & 82.23 $\pm$ 2.54 \\
                       & Reddit      & 80.03 $\pm$ 0.36 & 79.83 $\pm$ 0.31 & 79.55 $\pm$ 0.20 & \underline{81.22 $\pm$ 0.61} & 80.82 $\pm$ 0.45 & 73.59 $\pm$ 0.00 & 77.14 $\pm$ 0.16 & 78.44 $\pm$ 0.18 & \textbf{81.57 $\pm$ 0.67} \\
                       & MOOC        & 78.94 $\pm$ 1.25 & 75.60 $\pm$ 1.12 & 82.19 $\pm$ 0.62 & \textbf{87.06 $\pm$ 1.93} & 74.05 $\pm$ 0.95 & 60.71 $\pm$ 0.00 & 77.06 $\pm$ 0.41 & 77.77 $\pm$ 0.92 & \underline{85.85 $\pm$ 0.66} \\
                       & LastFM      & 74.35 $\pm$ 3.81 & 74.92 $\pm$ 2.46 & 71.59 $\pm$ 0.24 & \underline{76.87 $\pm$ 4.64} & 69.86 $\pm$ 0.43 & 73.03 $\pm$ 0.00 & 59.30 $\pm$ 2.31 & 72.47 $\pm$ 0.49 & \textbf{81.57 $\pm$ 0.48} \\
                       & Enron       & 69.85 $\pm$ 2.70 & 71.19 $\pm$ 2.76 & 64.07 $\pm$ 1.05 & 73.91 $\pm$ 1.76 & 64.73 $\pm$ 0.36 & \underline{76.53 $\pm$ 0.00} & 70.66 $\pm$ 0.39 & \textbf{77.98 $\pm$ 0.92} & 75.63 $\pm$ 0.73 \\
                       & Social Evo. & 87.44 $\pm$ 6.78 & 93.29 $\pm$ 0.43 & \underline{95.01 $\pm$ 0.44} & 94.45 $\pm$ 0.56 & 85.53 $\pm$ 0.38 & 80.57 $\pm$ 0.00 & 94.74 $\pm$ 0.31 & 94.93 $\pm$ 0.31 & \textbf{97.38 $\pm$ 0.14} \\
                       & UCI         & 75.24 $\pm$ 5.80 & 55.10 $\pm$ 3.14 & 68.27 $\pm$ 1.37 & 80.43 $\pm$ 2.12 & 65.30 $\pm$ 0.43 & 65.50 $\pm$ 0.00 & 80.25 $\pm$ 2.74 & \textbf{84.11 $\pm$ 1.35} & \underline{82.17 $\pm$ 0.82} \\
                       & Flights     & 66.48 $\pm$ 2.59 & 67.61 $\pm$ 0.99 & \textbf{72.38 $\pm$ 0.18} & 66.70 $\pm$ 1.64 & 64.72 $\pm$ 0.97 & 70.53 $\pm$ 0.00 & 70.68 $\pm$ 0.24 & \underline{71.47 $\pm$ 0.26} & 66.59 $\pm$ 0.49 \\
                       & Can. Parl.  & 51.79 $\pm$ 0.63 & 63.31 $\pm$ 1.23 & 67.13 $\pm$ 0.84 & 68.42 $\pm$ 3.07 & 66.53 $\pm$ 2.77 & 63.84 $\pm$ 0.00 & 65.93 $\pm$ 3.00 & \underline{74.34 $\pm$ 0.87} & \textbf{97.00 $\pm$ 0.31} \\
                       & US Legis.   & 51.71 $\pm$ 5.76 & \textbf{86.88 $\pm$ 2.25} & 62.14 $\pm$ 6.60 & 74.00 $\pm$ 7.57 & 68.82 $\pm$ 8.23 & 63.22 $\pm$ 0.00 & 80.53 $\pm$ 3.95 & 81.65 $\pm$ 1.02 & \underline{85.30 $\pm$ 3.88} \\
                       & UN Trade    & 61.39 $\pm$ 1.83 & 59.19 $\pm$ 1.07 & 55.74 $\pm$ 0.91 & 58.44 $\pm$ 5.51 & 55.71 $\pm$ 0.38 & \textbf{81.32 $\pm$ 0.00} & 55.90 $\pm$ 1.17 & 57.05 $\pm$ 1.22 & \underline{64.41 $\pm$ 1.40} \\
                       & UN Vote     & \underline{70.02 $\pm$ 0.81} & 69.30 $\pm$ 1.12 & 52.96 $\pm$ 2.14 & 69.37 $\pm$ 3.93 & 51.26 $\pm$ 0.04 & \textbf{84.89 $\pm$ 0.00} & 52.30 $\pm$ 2.35 & 51.20 $\pm$ 1.60 & 60.84 $\pm$ 1.58 \\
                       & Contact     & 95.31 $\pm$ 2.13 & \underline{96.39 $\pm$ 0.20} & 96.05 $\pm$ 0.52 & 93.05 $\pm$ 2.35 & 84.16 $\pm$ 0.49 & 88.81 $\pm$ 0.00 & 93.86 $\pm$ 0.21 & 93.36 $\pm$ 0.41 & \textbf{97.57 $\pm$ 0.06} \\ \cline{2-11} 
                       & Avg. Rank   & 5.46         & 5.08         & 5.08         & \underline{3.85}         & 7.54         & 5.92         & 5.46         & 4.00         & \textbf{2.62}         \\ \hline
\multirow{14}{*}{ind}  & Wikipedia   & 75.65 $\pm$ 0.79 & 70.21 $\pm$ 1.58 & \underline{87.00 $\pm$ 0.16} & 85.62 $\pm$ 0.44 & 74.06 $\pm$ 2.62 & 80.63 $\pm$ 0.00 & 86.76 $\pm$ 0.72 & \textbf{88.59 $\pm$ 0.17} & 78.29 $\pm$ 5.38 \\
                       & Reddit      & 86.98 $\pm$ 0.16 & 86.30 $\pm$ 0.26 & 89.59 $\pm$ 0.24 & 88.10 $\pm$ 0.24 & \textbf{91.67 $\pm$ 0.24} & 85.48 $\pm$ 0.00 & 87.45 $\pm$ 0.29 & 85.26 $\pm$ 0.11 & \underline{91.11 $\pm$ 0.40} \\
                       & MOOC        & 65.23 $\pm$ 2.19 & 61.66 $\pm$ 0.95 & 75.95 $\pm$ 0.64 & \underline{77.50 $\pm$ 2.91} & 73.51 $\pm$ 0.94 & 49.43 $\pm$ 0.00 & 74.65 $\pm$ 0.54 & 74.27 $\pm$ 0.92 & \textbf{81.24 $\pm$ 0.69} \\
                       & LastFM      & 62.67 $\pm$ 4.49 & 64.41 $\pm$ 2.70 & 71.13 $\pm$ 0.17 & 65.95 $\pm$ 5.98 & 67.48 $\pm$ 0.77 & \textbf{75.49 $\pm$ 0.00} & 58.21 $\pm$ 0.89 & 68.12 $\pm$ 0.33 & \underline{73.97 $\pm$ 0.50} \\
                       & Enron       & 68.96 $\pm$ 0.98 & 67.79 $\pm$ 1.53 & 63.94 $\pm$ 1.36 & 70.89 $\pm$ 2.72 & \underline{75.15 $\pm$ 0.58} & 73.89 $\pm$ 0.00 & 71.29 $\pm$ 0.32 & 75.01 $\pm$ 0.79 & \textbf{77.41 $\pm$ 0.89} \\
                       & Social Evo. & 89.82 $\pm$ 4.11 & 93.28 $\pm$ 0.48 & 94.84 $\pm$ 0.44 & \underline{95.13 $\pm$ 0.56} & 88.32 $\pm$ 0.27 & 83.69 $\pm$ 0.00 & 94.90 $\pm$ 0.36 & 94.72 $\pm$ 0.33 & \textbf{97.68 $\pm$ 0.10} \\
                       & UCI         & 65.99 $\pm$ 1.40 & 54.79 $\pm$ 1.76 & 68.67 $\pm$ 0.84 & 70.94 $\pm$ 0.71 & 64.61 $\pm$ 0.48 & 57.43 $\pm$ 0.00 & \underline{76.01 $\pm$ 1.11} & \textbf{80.10 $\pm$ 0.51} & 72.25 $\pm$ 1.71 \\
                       & Flights     & 69.07 $\pm$ 4.02 & 70.57 $\pm$ 1.82 & \underline{75.48 $\pm$ 0.26} & 71.09 $\pm$ 2.72 & 69.18 $\pm$ 1.52 & \textbf{81.08 $\pm$ 0.00} & 74.62 $\pm$ 0.18 & 74.87 $\pm$ 0.21 & 70.92 $\pm$ 1.78 \\
                       & Can. Parl.  & 48.42 $\pm$ 0.66 & 58.61 $\pm$ 0.86 & 68.82 $\pm$ 1.21 & 65.34 $\pm$ 2.87 & 67.75 $\pm$ 1.00 & 62.16 $\pm$ 0.00 & 65.85 $\pm$ 1.75 & \underline{69.48 $\pm$ 0.63} & \textbf{95.44 $\pm$ 0.57} \\
                       & US Legis.   & 50.27 $\pm$ 5.13 & \textbf{83.44 $\pm$ 1.16} & 61.91 $\pm$ 5.82 & 67.57 $\pm$ 6.47 & 65.81 $\pm$ 8.52 & 64.74 $\pm$ 0.00 & 78.15 $\pm$ 3.34 & 79.63 $\pm$ 0.84 & \underline{81.25 $\pm$ 3.62} \\
                       & UN Trade    & 60.42 $\pm$ 1.48 & 60.19 $\pm$ 1.24 & 60.61 $\pm$ 1.24 & 61.04 $\pm$ 6.01 & \underline{62.54 $\pm$ 0.67} & \textbf{72.97 $\pm$ 0.00} & 61.06 $\pm$ 1.74 & 60.15 $\pm$ 1.29 & 55.79 $\pm$ 1.02 \\
                       & UN Vote     & \textbf{67.79 $\pm$ 1.46} & 67.53 $\pm$ 1.98 & 52.89 $\pm$ 1.61 & \underline{67.63 $\pm$ 2.67} & 52.19 $\pm$ 0.34 & 66.30 $\pm$ 0.00 & 50.62 $\pm$ 0.82 & 51.60 $\pm$ 0.73 & 51.91 $\pm$ 0.84 \\
                       & Contact     & 93.43 $\pm$ 1.78 & 94.18 $\pm$ 0.10 & \underline{94.35 $\pm$ 0.48} & 90.18 $\pm$ 3.28 & 89.31 $\pm$ 0.27 & 85.20 $\pm$ 0.00 & 91.35 $\pm$ 0.21 & 90.87 $\pm$ 0.35 & \textbf{94.75 $\pm$ 0.28} \\ \cline{2-11} 
                       & Avg. Rank   & 6.62         & 6.38         & \underline{4.15}         & 4.38         & 5.46         & 5.62         & 4.69         & 4.46         & \textbf{3.23}         \\ \hline
\end{tabular}
}
}
\end{table}
We report the performance of different methods on the AP metric for transductive dynamic link prediction with three negative sampling strategies in \tabref{tab:average_precision_transductive_dynamic_link_prediction}. The best and second-best results are emphasized by \textbf{bold} and \underline{underlined} fonts. Note that the results are multiplied by 100 for a better display layout. Please refer to \secref{section-appendix-numerical-auc-roc-performance-transductive-dynamic-link-prediction} and \secref{section-appendix-numerical-performance-inductive-dynamic-link-prediction} for the results of AP for inductive dynamic link prediction as well as AUC-ROC for transductive and inductive dynamic link prediction tasks. Since EdgeBank can be only evaluated for transductive dynamic link prediction, we do not show its performance under the inductive setting. 
From the results, we have two main observations. 

Firstly, DyGFormer usually outperforms baselines and achieves an average rank of 2.49/2.69 on AP/AUC-ROC for transductive and 2.69/2.56 for inductive dynamic link prediction across three negative sampling strategies. This is because: (\romannumeral1) The neighbor co-occurrence encoding scheme helps DyGFormer exploit correlations between the source node and destination node, which are often predictive for future links (see \secref{section-verify-motivation-neighbor-co-occurrence-encoding}). (\romannumeral2) The patching technique allows DyGFormer to access longer histories and capture long-term temporal dependencies (see \secref{section-5-investigation_patch_technique}). In \tabref{tab:num_neighbors_configuration} in \secref{section-appendix-configurations}, the input sequence lengths of DyGFormer are much longer than those of baselines on several datasets, indicating that it can utilize longer sequences better. We also observe the varying results of DyGFormer across different negative sampling strategies and give an analysis in \secref{section-varying-performance-DyGFormer-negative-sampling-strategies}.

Secondly, some of our findings of baselines differ from previous reports. For instance, the performance of some baselines can be significantly improved by properly setting some hyperparameters. Additionally, some methods would obtain worse results after we fix the problems or make adaptions in their implementations. More explanations can be found in \secref{section-appendix-inconsistent-observations}. These observations highlight the importance of rigorously evaluating different methods by a unified library and verify the necessity of introducing DyGLib to facilitate the development of dynamic graph learning.

We also report the results of dynamic node classification in \tabref{tab:auc_roc_dynamic_node_classification} in \secref{section-appendix-node-clasification}. We observe that DyGFormer obtains better performance than most baselines and achieves an impressive average rank of 2.50 among them, demonstrating the superiority of DyGFormer once again. 

\begin{figure}[!htbp] 
\centering
\begin{minipage}[c]{0.39\linewidth} 
\centering 
\captionof{table}{AP for TCL with NCoE.} 
\label{tab:tcl_generalizability_neighbor_co_occurrence_encoding} 
\setlength{\tabcolsep}{0.7mm}
{
\begin{tabular}{c|ccc}
\hline
Datasets    & TCL    & w/ NCoE & Improv. \\ \hline
Wikipedia   & 96.47 & 99.09      & 2.72\%  \\
Reddit      & 97.53 & 99.04      & 1.55\%  \\
MOOC        & 82.38 & 86.92      & 5.51\%  \\
LastFM      & 67.27 & 84.02      & 24.90\% \\
Enron       & 79.70 & 90.18      & 13.15\% \\
Social Evo. & 93.13 & 94.06      & 1.00\%  \\
UCI         & 89.57 & 94.69      & 5.72\%  \\
Flights     & 91.23 & 97.71      & 7.10\%  \\
Can. Parl.  & 68.67 & 69.34      & 0.98\%  \\
US Legis.   & 69.59 & 69.47      & -0.17\% \\
UN Trade    & 62.21 & 63.46      & 2.01\%  \\
UN Vote     & 51.90 & 51.52      & -0.73\% \\
Contact     & 92.44 & 97.98      & 5.99\%  \\ \hline
\end{tabular}
\vspace{-13pt}
}
\end{minipage} 
\hfill 
\begin{minipage}[c]{0.6\linewidth} 
\centering 
    \vspace{20pt}
    \includegraphics[width=1.0\columnwidth]{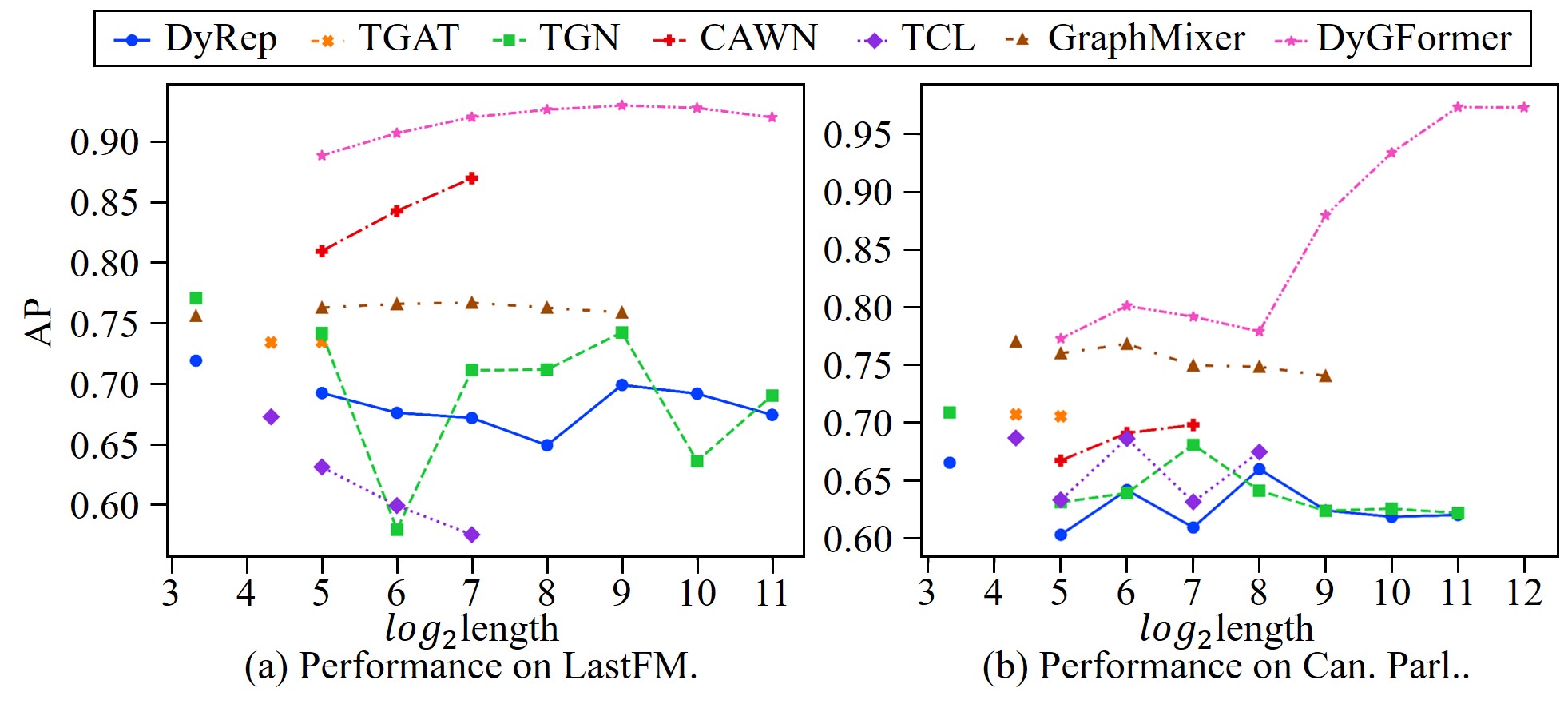}
    \captionof{figure}{Performance of different methods on LastFM and Can. Parl. with varying input lengths.}
    \label{fig:methods_varying_input_lengths}
\end{minipage} 
\end{figure}

\subsection{Generalizability of Neighbor Co-occurrence Encoding Scheme}\label{section-5-node_cooccurrence_generalizability}
Our \textbf{N}eighbor \textbf{Co}-occurrence \textbf{E}ncoding scheme (NCoE) is versatile and can be easily integrated with dynamic graph learning methods based on sequential models. Hence, we incorporate NCoE with TCL and GraphMixer and show their performance in \tabref{tab:tcl_generalizability_neighbor_co_occurrence_encoding} and \tabref{tab:complete_generalizability_neighbor_co_occurrence_encoding} in \secref{section-appendix-node-cooccurrence-generalizability}. We find TCL and GraphMixer usually yield better results with NCoE, achieving an average improvement of 5.36\% and 1.86\% over all datasets. This verifies the effectiveness and versatility of the neighbor co-occurrence encoding, and highlights the importance of capturing correlations between nodes. Also, as TCL and DyGFormer are built upon Transformer, TCL w/ NCoE can achieve similar results with DyGFormer on datasets that enjoy shorter input sequences (in which cases the patching technique in DyGFormer contributes little). However, when datasets exhibit more obvious long-term temporal dependencies (e.g., LastFM, Can. Parl.), the performance gaps become more significant.

\subsection{Advantages of Patching Technique}\label{section-5-investigation_patch_technique}
We validate the advantages of our patching technique in preserving the local temporal proximities and reducing the computational complexity, which helps DyGFormer effectively and efficiently utilize longer histories. We conduct experiments on LastFM and Can. Parl. since they can benefit from longer historical records. For baselines, we sample more neighbors or perform more causal anonymous walks (starting from 32) to make them access longer histories. The results are depicted in \figref{fig:methods_varying_input_lengths}, where the x-axis is represented by a logarithmic scale with base 2. We also plot the performance of baselines with the optimal length by unconnected points based on \tabref{tab:num_neighbors_configuration} in \secref{section-appendix-configurations}. Note that the results of some baselines are incomplete since they raise the out-of-memory error when the lengths are longer. For example, TGAT is only computationally feasible when extending the input length to 32, resulting in two discrete points with lengths 20 (the optimal length) and 32. 

From \figref{fig:methods_varying_input_lengths}, we conclude that: (\romannumeral1) most of the baselines perform worse when the input lengths become longer, indicating they lack the ability to capture long-term temporal dependencies; (\romannumeral2) the baselines usually encounter expensive computational costs when computing on longer histories. Although memory network-based methods (i.e., DyRep and TGN) can handle longer histories with affordable computational costs, they cannot benefit from longer histories due to the potential issues of vanishing or exploding gradients; (\romannumeral3) DyGFormer consistently achieves gains from longer sequences, demonstrating the advantages of the patching technique in leveraging longer histories. 

We also compare the running time and memory usage of DyGFormer with and without the patching technique during the training process. The results are shown in \tabref{tab:comparison_training_time_memory_usage_with_without_patching} in \secref{section-appendix-comparison-training-time-memory-usage}. We could observe that the patching technique efficiently reduces model training costs in both time and space, allowing DyGFormer to access longer histories. As the input sequence length increases, the reductions become more significant. Moreover, with the patching technique, we find that DyGFormer achieves an average improvement of 0.31\% and 0.74\% in performance on LastFM and Can. Parl. than DyGFormer without patching. This observation further demonstrates the advantage of our patching technique in leveraging the local temporal proximities for better results.

\subsection{Verification of the Motivation of Neighbor Co-occurrence Encoding Scheme}\label{section-verify-motivation-neighbor-co-occurrence-encoding}
To verify the motivation of NCoE (i.e., nodes with more common historical neighbors tend to interact in the future), we compare the performance of DyGFormer and DyGFormer without NCoE. We choose 0.5 as the threshold and use TP, TN, FN, and FP to denote True/False Positive/Negative. Common Neighbor Ratio (CNR) is defined as the ratio of common neighbors in source node $u$’s sequence $S_u$ and destination node $v$’s sequence $S_v$, i.e., $|S_u \cap S_v|/|S_u \cup S_v|$. We focus on links whose predictions of DyGFormer w/o NCoE are changed by DyGFormer (i.e., FN→TP, FP→TN, TP→FN, and TN→FP). We define Changed Link Ratio (CLR) as the ratio of the changed links to their original set, which is respectively computed by $|$FN→TP$|/|$FN$|$, $|$FP→TN$|/|$FP$|$, $|$TP→FN$|/|$TP$|$, and $|$TN→FP$|/|$TN$|$. If NCoE is helpful, DyGFormer will revise more wrong predictions (more FN→TP and FP→TN) and make fewer incorrect changes (fewer TP→FN and TN→FP). We report CLR and average CNR of links in the above sets on five typical datasets in \tabref{tab:CLR_CNR_modification}.

\begin{table}[!htbp]
\centering
\caption{CLR and CNR of changes made by DyGFormer.}
\label{tab:CLR_CNR_modification}
\resizebox{0.97\textwidth}{!}
{
\setlength{\tabcolsep}{0.9mm}
{
\begin{tabular}{c|cccc|cccc}
\hline
\multirow{2}{*}{Datasets} & \multicolumn{4}{c|}{CLR (\%)}                & \multicolumn{4}{c}{CNR (\%)}                 \\ \cline{2-9} 
                          & FN→TP & FP→TN & TP→FN & TN→FP & FN→TP & FP→TN & TP→FN & TN→FP \\ \hline
Wikipedia                 & 68.36     & 72.73     & 1.68      & 1.69      & 18.16     & 0.01      & 0.10      & 2.49      \\
UCI                       & 71.45     & 94.11     & 7.29      & 1.82      & 19.08     & 2.49      & 3.35      & 13.02     \\
Flights                   & 83.66     & 83.83     & 1.73      & 2.11      & 37.09     & 2.28      & 7.06      & 20.28     \\
US Legis.                 & 31.63     & 23.67     & 6.63      & 1.59      & 69.92     & 62.13     & 61.14     & 63.80     \\
UN Vote                   & 44.02     & 36.46     & 28.95     & 30.53     & 78.57     & 81.39     & 80.86     & 77.02     \\ \hline
\end{tabular}
}
}
\end{table}

We find NCoE effectively helps DyGFormer rectify wrong predictions of DyGFormer w/o NCoE on datasets with \textit{significantly higher CNR of positive links than negative ones}, which happens with most datasets. Concretely, for Wikipedia, UCI, and Flights, their CNRs of FN→TP are much higher than FP→TN (e.g., 37.09\% vs. 2.28\% on Flights) and DyGFormer revises most wrong predictions of DyGFormer w/o NCoE (e.g., 83.66\% for positive links in FN and 83.83\% for negative links in FP on Flights). Corrections made by our encoding scheme are less obvious on datasets whose \textit{CNRs between positive and negative links are similar}, which occurs in only 2 of 13 datasets. For US Legis. and UN Vote, their CNRs between FN and FP are analogous (e.g., 69.92\% vs. 62.13\% on US Legis.), weakening the advantage of our neighbor co-occurrence encoding scheme (e.g., only 31.63\%/23.67\% of positive/negative links are corrected in FN/FP on US Legis.). Therefore, we conclude that the neighbor co-occurrence encoding scheme helps DyGFormer capture common historical neighbors in $S_u$ and $S_v$, and bring better results in most cases. 

\subsection{When Will DyGFormer Be a Good Choice?}\label{section-when-will-dygformer-good}
Note that DyGFormer is superior to baselines by 1) exploring the source and destination nodes’ correlations from their historical sequences by neighbor co-occurrence encoding scheme; 2) using the patching technique to attend longer histories. Thus, DyGFormer tends to perform better on datasets that favor these two designs. 
We define Link Ratio (LR) as the ratio of links in their corresponding positive or negative set, which can be computed by TP$/$(TP+FN), TN$/$(TN+FP), FN$/$(TP+FN), and FP$/$(TN+FP). As a method with more TP and TN (i.e., fewer FN and FP) is better, we report the results of LR and average CNR of links in TP and TN on five typical datasets in \tabref{tab:LR_CNR_TP_TN}.

\begin{table}[!htbp]
\begin{minipage}{0.45\linewidth} 
\centering 
\caption{LR and CNR of TP and TN with random negative sampling strategy.}
\label{tab:LR_CNR_TP_TN}
\setlength{\tabcolsep}{1.0mm}
{
\begin{tabular}{c|cc|cc}
\hline
\multirow{2}{*}{Datasets} & \multicolumn{2}{c|}{LR (\%)} & \multicolumn{2}{c}{CNR (\%)} \\ \cline{2-5} 
                          & TP       & TN       & TP       & TN       \\ \hline
Wikipedia                 & 92.74      & 97.19      & 59.09      & 0.01       \\
UCI                       & 82.70      & 96.77      & 28.03      & 1.45       \\
Flights                   & 96.13      & 95.33      & 47.58      & 1.40     \\  
US Legis.                 & 78.95      & 56.83      & 75.18      & 53.98      \\
UN Vote                   & 65.18      & 45.43      & 56.24      & 76.02      \\ \hline
\end{tabular}
}
\end{minipage} 
\begin{minipage}{0.55\linewidth} 
\centering 
\caption{LR and CNR of FP under random, historical, and inductive negative sampling strategy.}
\label{tab:rnd_hist_ind_fp}
\setlength{\tabcolsep}{1.0mm}
{
\begin{tabular}{c|ccc|ccc}
\hline
\multirow{2}{*}{Datasets} & \multicolumn{3}{c|}{LR(\%)} & \multicolumn{3}{c}{CNR(\%)} \\ \cline{2-7} 
                          & rnd      & hist     & ind     & rnd      & hist     & ind     \\ \hline
Wikipedia                 & 2.81     & 89.28    & 94.53   & 0.02     & 14.00    & 11.66   \\
UCI                       & 3.23     & 64.93    & 76.42   & 9.98     & 12.22    & 13.81   \\
Flights                   & 4.67     & 94.52    & 92.94   & 0.01     & 35.62    & 30.29   \\
US Legis.                 & 43.17    & 17.31    & 21.21   & 79.40    & 87.51    & 75.51   \\
UN Vote                   & 54.57    & 39.90    & 52.92   & 79.60    & 75.53    & 79.15   \\ \hline
\end{tabular}
}
\end{minipage} 
\end{table}

We observe when \textit{CNR of TP is significantly higher than CNR of TN in the datasets, DyGFormer often outperforms baselines} (most datasets satisfy this property). For Wikipedia, UCI, and Flights, their CNRs of TP are much higher than those of TN (e.g., 59.09\% vs. 0.01\% on Wikipedia). Such a characteristic matches the motivation of our neighbor co-occurrence encoding scheme, enabling DyGFormer to correctly predict most links (e.g., 92.74\% of positive links and 97.19\% of negative links are properly predicted on Wikipedia). Moreover, as LastFM and Can. Parl. can gain from longer histories (see \figref{fig:methods_varying_input_lengths} and \tabref{tab:num_neighbors_configuration} in \secref{section-appendix-configurations}), DyGFormer is significantly better than baselines on these two datasets. When \textit{CNRs of TP and TN are less distinguishable in the datasets, DyGFormer may perform worse} (only 2 out of 13 datasets show this property). For US Legis., the CNRs between TP and TN are close (i.e., 75.18\% vs. 53.98\%), making DyGFormer worse than memory-based baselines (i.e., JODIE, DyRep, and TGN). For UN Vote, its CNR of TP is even lower than that of TN (i.e., 56.24\% vs. 76.02\%), which is opposite to our motivation, leading to poor results of DyGFormer than a few baselines. Since these two datasets cannot obviously gain from longer sequences either (see \tabref{tab:num_neighbors_configuration} in \secref{section-appendix-configurations}), DyGFormer obtains worse results on them. Thus, we conclude that for datasets with much higher CNR of TP than CNR of TN or datasets that can benefit from longer histories, DyGFormer is a good choice. Otherwise, we may need to try other methods.

\subsection{Why do DyGFormer's Performance Vary across Different Negative Sampling Strategies?}\label{section-varying-performance-DyGFormer-negative-sampling-strategies}
Compared with the random (rnd) strategy, historical (hist) and inductive (ind) strategies will sample previous links as negative ones. This makes previous positive links negative, which may hurt the performance DyGFormer since the motivation of our neighbor co-occurrence encoding scheme is violated. As positive links are identical among rnd, hist, and ind, we compute LR and the average CNR of links in FP and show results in \tabref{tab:rnd_hist_ind_fp}.

We find \textit{when hist or ind causes several magnitudes higher CNR of FP than rnd in the datasets, DyGFormer drops sharply}. For Wikipedia, UCI, and Flights, the CNRs of FP with hist/ind are much higher than rnd (e.g., 14.00\%/11.66\% vs. 0.02\% on Wikipedia). This misleads DyGFormer to predict negative links as positive and causes drops (e.g., 89.28\%/94.53\% of negative links are incorrectly predicted with hist/ind on Wikipedia, while only 2.81\% are wrong with rnd). We also note the drops in UCI are milder since the changes in CNR caused by hist or ind vs. rnd are less obvious than changes in Wikipedia and Flights. \textit{When changes in CNR of FP caused by hist or ind are not obvious in the datasets, DyGFormer is less affected}. Since hist/ind makes little changes in CNRs of FP on US Legis., we find it ranks second with hist/ind, which may indicate DyGFormer is less influenced by the neighbor co-occurrence encoding scheme and generalizes well to various negative sampling strategies. For UN Vote, although its CNRs of FP are not affected by hist and ind either, DyGFormer still performs worse due to its inferior performance with rnd. Hence, we deduce that our neighbor co-occurrence encoding may be sometimes fragile to various negative sampling strategies if its motivation is violated, leading to the varying performance of DyGFormer.

\subsection{Ablation Study}\label{section-5-ablation_study}
Finally, we validate the effectiveness of the neighbor co-occurrence encoding, time encoding, and mixing of the sequence of source node and destination node in DyGFormer. From \figref{fig:ablation_study} in \secref{section-appendix-ablation-study}, we observe that DyGFormer obtains the best performance when using all the components, and the results would be worse when any component is removed. This illustrates the necessity of each design in DyGFormer. Please refer to \secref{section-appendix-ablation-study} for detailed implementations and discussions.

\section{Conclusion}
\label{section-6}
In this paper, we proposed a new Transformer-based architecture (DyGFormer) and a unified library (DyGLib) to foster the development of dynamic graph learning. DyGFormer differs from previous methods in (\romannumeral1) a neighbor co-occurrence encoding scheme to exploit the correlations of nodes in each interaction; and (\romannumeral2) a patching technique to help the model capture long-term temporal dependencies. DyGLib served as a toolkit for reproducible, scalable, and credible continuous-time dynamic graph learning with standard training pipelines, extensible coding interfaces, and comprehensive evaluating protocols. We hope our work can provide new perspectives on designing new dynamic graph learning frameworks and encourage more researchers to dive into this field. In the future, we will continue to enrich DyGLib by incorporating the recently released datasets and state-of-the-art models.

\section*{Acknowledgments and Disclosure of Funding}
This work was supported by the National Natural Science Foundation of China (No. 62272023 and 51991395) and the Fundamental Research Funds for the Central Universities (No. YWF-23-L-1203).

\bibliographystyle{plain}
\bibliography{reference.bib}

\clearpage
\appendix
\label{section-appendix}

\section{Additional Discussions and Details}
\subsection{Limitations}
One potential limitation of DyGFormer lies in the ignorance of high-order relationships between nodes since it solely learns from the first-hop interactions of nodes. In certain scenarios where nodes' high-order relationships are essential, DyGFormer may be suboptimal compared with baselines that learn the higher-order interactions. However, trivially feeding the multi-hop neighbors of nodes into DyGFormer would incur expensive computational costs. It is promising to design more efficient and effective frameworks to model nodes' high-order relationships for dynamic graph learning.

Another potential limitation is the sensitivity of the neighbor co-occurrence encoding scheme against different negative sampling strategies (discussed in \secref{section-varying-performance-DyGFormer-negative-sampling-strategies}). When the assumption of our neighbor co-occurrence encoding scheme is violated, its performance may drop drastically in some cases. It is an insightful direction to design more robust encoding schemes to tackle this issue.

\subsection{Licenses}
All the used codes and datasets are publicly available and permit usage for research purposes under either MIT License or Apache License 2.0.

\subsection{Overall Procedure of DyGLib}\label{section-appendix-DyGLib-procedure}
\begin{figure}[!ht]
    \centering
    \includegraphics[scale=0.5]{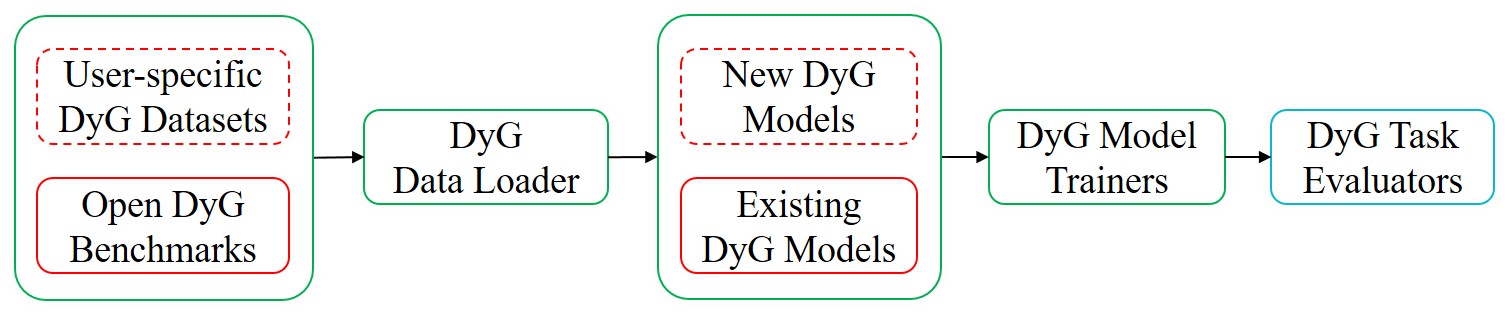}
    \caption{DyGLib is equipped with \textcolor[RGB]{0,176,80}{standard training pipelines}, \textcolor[RGB]{255,0,0}{extensible coding interfaces}, and \textcolor[RGB]{9,182,205}{comprehensive evaluating protocols}. DyG denotes the abbreviation of Dynamic Graph.}
    \label{fig:procedure_DyGLib}
\end{figure}

\figref{fig:procedure_DyGLib} shows the overall procedure of DyGLib.

\section{Detailed Experimental Settings}
\subsection{Descriptions of Datasets}\label{section-appendix-descriptions_datasets}
We use thirteen datasets collected by \cite{poursafaei2022towards} in the experiments, which are publicly available\footnote{\url{https://zenodo.org/record/7213796\#.Y1cO6y8r30o}}:
\begin{itemize}
    \item  \textbf{Wikipedia} is a bipartite interaction graph that contains the edits on Wikipedia pages over a month. Nodes represent users and pages, and links denote the editing behaviors with timestamps. Each link is associated with a 172-dimensional Linguistic Inquiry and Word Count (LIWC) feature \cite{pennebaker2001linguistic}. This dataset additionally contains dynamic labels that indicate whether users are temporarily banned from editing.

    \item  \textbf{Reddit} is bipartite and records the posts of users under subreddits during one month. Users and subreddits are nodes, and links are the timestamped posting requests. Each link has a 172-dimensional LIWC feature. This dataset also includes dynamic labels representing whether users are banned from posting.

    \item  \textbf{MOOC} is a bipartite interaction network of online sources, where nodes are students and course content units (e.g., videos and problem sets). Each link denotes a student's access behavior to a specific content unit and is assigned with a 4-dimensional feature.

    \item  \textbf{LastFM} is bipartite and consists of the information about which songs were listened to by which users over one month. Users and songs are nodes, and links denote the listening behaviors of users.

    \item  \textbf{Enron} records the email communications between employees of the ENRON energy corporation over three years.

    \item  \textbf{Social Evo.} is a mobile phone proximity network that monitors the daily activities of an entire undergraduate dormitory for a period of eight months, where each link has a 2-dimensional feature.

    \item  \textbf{UCI} is an online communication network, where nodes are university students and links are messages posted by students.
    
    \item  \textbf{Flights} is a dynamic flight network that displays the development of air traffic during the COVID-19 pandemic. Airports are represented by nodes and the tracked flights are denoted as links. Each link is associated with a weight, indicating the number of flights between two airports in a day.

    \item  \textbf{Can. Parl.} is a dynamic political network that records the interactions between Canadian Members of Parliament (MPs) from 2006 to 2019. Each node represents an MP from an electoral district and a link is created when two MPs both vote "yes" on a bill. The weight of each link refers to the number of times that one MP voted “yes” for another MP in a year. 
    
    \item  \textbf{US Legis.} is a senate co-sponsorship network that tracks social interactions between legislators in the US Senate. The weight of each link specifies the number of times two congresspersons have co-sponsored a bill in a given congress.

    \item  \textbf{UN Trade} contains the food and agriculture trade between 181 nations for more than 30 years. The weight of each link indicates the total sum of normalized agriculture import or export values between two particular countries.

    \item  \textbf{UN Vote} records roll-call votes in the United Nations General Assembly. If two nations both voted "yes" to an item, the weight of the link between them is increased by one.

    \item  \textbf{Contact} describes how the physical proximity evolves among about 700 university students over a month. Each student has a unique identifier and links denote that they are within close proximity to each other. Each link is associated with a weight, revealing the physical proximity between students.
\end{itemize}
We show the statistics of the datasets in \tabref{tab:data_statistics}, where \#N\&L Feat stands for the dimensions of node and link features. We notice a slight difference between the statistics of the Contact dataset reported in \cite{poursafaei2022towards} (which has 694 nodes and 2,426,280 links) and our own calculations, although both of them are computed based on the released dataset by \cite{poursafaei2022towards}. We ultimately report our statistics of the Contact dataset in this paper.

\begin{table}[!htbp]
\centering
\caption{Statistics of the datasets.}
\label{tab:data_statistics}
\resizebox{1.01\textwidth}{!}
{
\setlength{\tabcolsep}{0.45mm}
{
\begin{tabular}{c|cccccccc}
\hline
Datasets    & Domains     & \#Nodes & \#Links   & \#N\&L Feat & Bipartite & Duration  & Unique Steps & Time Granularity    \\ \hline
Wikipedia   & Social      & 9,227  & 157,474   & -- \& 172                & True      & 1 month    &  152,757      &  Unix timestamps   \\
Reddit      & Social      & 10,984 & 672,447   & -- \& 172                & True      & 1 month    &  669,065      &  Unix timestamps          \\
MOOC        & Interaction & 7,144  & 411,749   & -- \& 4                  & True      & 17 months    &  345,600      & Unix timestamps         \\
LastFM      & Interaction & 1,980  & 1,293,103 & -- \& --                 & True      & 1 month    &   1,283,614     & Unix timestamps           \\
Enron       & Social      & 184    & 125,235   & -- \& --                 & False     & 3 years    &   22,632     &    Unix timestamps        \\
Social Evo. & Proximity   & 74     & 2,099,519 & -- \& 2                  & False     & 8 months    &  565,932      &  Unix timestamps         \\
UCI         & Social      & 1,899  & 59,835    & -- \& --                 & False     & 196 days    &  58,911      &  Unix timestamps         \\
Flights     & Transport   & 13,169 & 1,927,145 & -- \& 1                  & False     & 4 months    &  122      &   days        \\
Can. Parl.  & Politics    & 734    & 74,478    & -- \& 1                  & False     & 14 years    &   14     &   years        \\
US Legis.   & Politics    & 225    & 60,396    & -- \& 1                  & False     & 12 congresses    &   12     &  congresses    \\
UN Trade    & Economics   & 255    & 507,497   & -- \& 1                  & False     & 32 years    &    32    &   years        \\
UN Vote     & Politics    & 201    & 1,035,742 & -- \& 1                  & False     & 72 years    &    72    &     years      \\
Contact     & Proximity   & 692    & 2,426,279 & -- \& 1                  & False     & 1 month    &    8,064    &   5 minutes         \\ \hline
\end{tabular}
}
}
\end{table}

\subsection{Descriptions of Baselines}\label{section-appendix-descriptions_baselines}
We select the following eight baselines:
\begin{itemize}
    \item  \textbf{JODIE} is designed for temporal bipartite networks of user-item interactions. It employs two coupled recurrent neural networks to update the states of users and items. A projection operation is introduced to learn the future representation trajectory of each user/item \cite{DBLP:conf/kdd/KumarZL19}.
    
    \item  \textbf{DyRep} proposes a recurrent architecture to update node states upon each interaction. It also includes a temporal-attentive aggregation module to consider the temporally evolving structural information in dynamic graphs \cite{DBLP:conf/iclr/TrivediFBZ19}.

    \item  \textbf{TGAT} computes the node representation by aggregating features from each node's temporal-topological neighbors based on the self-attention mechanism. It is also equipped with a time encoding function for capturing temporal patterns \cite{DBLP:conf/iclr/XuRKKA20}.
    
    \item  \textbf{TGN} maintains an evolving memory for each node and updates this memory when the node is observed in an interaction, which is achieved by the message function, message aggregator, and memory updater. An embedding module is leveraged to generate the temporal representations of nodes \cite{DBLP:journals/corr/abs-2006-10637}.

    \item  \textbf{CAWN} first extracts multiple causal anonymous walks for each node, which can explore the causality of network dynamics and generate relative node identities. Then, it utilizes recurrent neural networks to encode each walk and aggregates these walks to obtain the final node representation \cite{DBLP:conf/iclr/WangCLL021}. 

    \item  \textbf{EdgeBank} is a pure memory-based approach without trainable parameters for transductive dynamic link prediction. It stores the observed interactions in the memory unit and updates the memory through various strategies. An interaction will be predicted as positive if it was retained in the memory, and negative otherwise \cite{poursafaei2022towards}. The publication presents two updating strategies, but their official implementations include two more\footnote{\url{https://github.com/fpour/DGB/blob/main/EdgeBank/link_pred/edge_bank_baseline.py}}. To be specific, the four variants of EdgeBank are: EdgeBank$_\infty$ with unlimited memory to store all the observed edges; EdgeBank$_\text{tw-ts}$ and EdgeBank$_\text{tw-re}$, which only remember edges within a fixed-size time window from the immediate past. The window size of EdgeBank$_\text{tw-ts}$ is set to the duration of the test split, while EdgeBank$_\text{tw-re}$ makes it related to the time intervals of repeated edges; EdgeBank$_\text{th}$ that retains edges with appearing counts higher than a threshold. We experiment with all four variants and report the best performance among them.
    
    \item  \textbf{TCL} first generates each node’s interaction sequence by performing a breadth-first search algorithm on the temporal dependency interaction sub-graph. Then, it presents a graph transformer that considers both graph topology and temporal information to learn node representations. It also incorporates a cross-attention operation for modeling the inter-dependencies between two interaction nodes \cite{DBLP:journals/corr/abs-2105-07944}.

    \item  \textbf{GraphMixer} shows that a fixed time encoding function performs better than the trainable version. It incorporates the fixed function into a link encoder based on MLP-Mixer \cite{DBLP:conf/nips/TolstikhinHKBZU21} to learn from temporal links. A node encoder with neighbor mean-pooing is employed to summarize node features \cite{cong2023do}.
\end{itemize}

\subsection{Some Problematic Implementations in Baselines}\label{section-appendix-issues-existing-methods}
JODIE, DyRep, and TGN models are based on memory networks, and their implementations have designed the raw messages to avoid information leakage. However, they fail to store the raw messages when saving models because the raw messages are maintained in a dictionary and thus cannot be saved as model parameters\footnote{\url{https://github.com/twitter-research/tgn/blob/2aa295a1f182137a6ad56328b43cb3d8223cae77/train_self_supervised.py\#L302}}. Our DyGLib has addressed this issue by additionally saving the correct raw messages when saving models. In the official implementation of CAWN, the encoding of each causal anonymous walk is represented by the embedding at the last position of the walk\footnote{\url{https://github.com/snap-stanford/CAW/blob/f994ff2b2c29778e6250b6a9928fd9943e0163f7/module.py\#L1069}}. However, some walks are padded to support mini-batch training in practice, making the last position of these padded walks meaningless. To get the correct encoding, it is necessary to use the actual length of each walk. Moreover, there are issues with the nodeedge2idx dictionary computed by the get\_ts2idx function\footnote{\url{https://github.com/snap-stanford/CAW/blob/f994ff2b2c29778e6250b6a9928fd9943e0163f7/graph.py\#L79}} if multiple interactions simultaneously occur at the last timestamp. This would lead to information leakage since the model can potentially access more interactions for a given interaction even if they happen at the same time. Our DyGLib has fixed these problems as well. 

\subsection{Some Inconsistent Observations with Previous Reports}\label{section-appendix-inconsistent-observations}
In the experiments, we find the behaviors of baselines are inconsistent with their previous reports in some cases. We provide detailed illustrations and attribute these phenomena to the following reasons.

\textbf{Suboptimal Settings of Hyperparameters}. Some important hyperparameters in the baselines are not sufficiently fine-tuned, such as the dropout rate, the number of sampled neighbors, the number of random walks, the length of input sequences, and the neighbor sampling strategies. In this paper, we perform the grid search to find the best settings of these hyperparameters 
and observe that the performance of many baselines can be significantly improved by properly setting certain hyperparameters. Take TGAT for transductive dynamic link prediction with the random negative sampling strategy as an example (see \tabref{tab:average_precision_transductive_dynamic_link_prediction}). Compared with the performance in \cite{poursafaei2022towards}, the results of AP are significantly improved on datasets like MOOC (from 0.61 to 0.86), LastFM (from 0.50 to 0.73), Enron (from 0.59 to 0.71), Social Evo. (from 0.76 to 0.93), Flights (from 0.89 to 0.94), and Contact (from 0.58 to 0.96). Similar improvements can also be found on JODIE, DyRep, and TGN on several datasets. 

\textbf{Usage of Problematic Implementations}. Some previous studies utilize problematic implementations in their experiments (illustrated in \secref{section-appendix-issues-existing-methods}), and the reported results may not be rigorous. Therefore, some methods would obtain worse results after we fix the problems in their implementations. Take CAWN for transductive dynamic link prediction with the random negative sampling strategy as an example (see \tabref{tab:average_precision_transductive_dynamic_link_prediction}). Compared with the results in \cite{poursafaei2022towards}, the performance on the AP metric of CAWN drops sharply on datasets like LastFM (from 0.98 to 0.87), Can. Parl. (from 0.94 to 0.70), US Legis. (from 0.97 to 0.71), UN Trade (from 0.97 to 0.65), and UN Vote (from 0.82 to 0.53). This is because \cite{poursafaei2022towards} uses the problematic implementations of CAWN to conduct experiments and the results will sometimes become worse after fixing the issues.

\textbf{Adaptions for Evaluations}. There also exist some differences between GraphMixer's results in the original paper and our work because we modify its implementation to fit our evaluations. The original GraphMixer could only be evaluated for transductive dynamic link prediction. In this work, we remove the one-hot encoding of nodes in GraphMixer to adapt it to the inductive setting, which may lead to decreased performance in certain situations. Take the performance of GraphMixer for transductive dynamic link prediction with the random negative sampling strategy as an example (see \tabref{tab:average_precision_transductive_dynamic_link_prediction}). Compared with the results in \cite{cong2023do}, the performance of GraphMixer slightly drops on datasets like Wikipedia (from 0.9985 to 0.9725) and Reddit (from 0.9993 to 0.9731).

\subsection{Configurations of Different Methods}\label{section-appendix-configurations}
We first present the official settings of baselines as well as the configurations of DyGFormer. These configurations remain unchanged across all the datasets.
\begin{itemize}
    \item  \textbf{JODIE}:
    \begin{itemize}
    \item Dimension of time encoding: 100
    \item Dimension of node memory: 172
    \item Dimension of output representation: 172
    \item Memory updater: vanilla recurrent neural network
    \end{itemize}

    \item  \textbf{DyRep}:
    \begin{itemize}
    \item Dimension of time encoding: 100
    \item Dimension of node memory: 172
    \item Dimension of output representation: 172
    \item Number of graph attention heads: 2
    \item Number of graph convolution layers: 1
    \item Memory updater: vanilla recurrent neural network
    \end{itemize}
    
    \item  \textbf{TGAT}:
    \begin{itemize}
    \item Dimension of time encoding: 100
    \item Dimension of output representation: 172
    \item Number of graph attention heads: 2
    \item Number of graph convolution layers: 2
    \end{itemize}
    
    \item  \textbf{TGN}:
    \begin{itemize}
    \item Dimension of time encoding: 100
    \item Dimension of node memory: 172
    \item Dimension of output representation: 172
    \item Number of graph attention heads: 2
    \item Number of graph convolution layers: 1
    \item Memory updater: gated recurrent unit \cite{DBLP:conf/ssst/ChoMBB14}
    \end{itemize}
    
    \item  \textbf{CAWN}:
    \begin{itemize}
    \item Dimension of time encoding: 100
    \item Dimension of position encoding: 172
    \item Dimension of output representation: 172
    \item Number of attention heads for encoding walks: 8
    \item Length of each walk (including the target node): 2
    \item Time scaling factor $\alpha$: 1e-6
    \end{itemize}
    
    \item  \textbf{TCL}:
    \begin{itemize}
    \item Dimension of time encoding: 100
    \item Dimension of depth encoding: 172
    \item Dimension of output representation: 172
    \item Number of attention heads: 2
    \item Number of Transformer layers: 2
    \end{itemize}

    \item  \textbf{GraphMixer}:
    \begin{itemize}
    \item Dimension of time encoding: 100
    \item Dimension of output representation: 172
    \item Number of MLP-Mixer layers: 2
    \item Time gap $T$: 2000
    \end{itemize}

    \item  \textbf{DyGFormer}:
    \begin{itemize}
    \item Dimension of time encoding $d_T$: 100
    \item Dimension of neighbor co-occurrence encoding $d_C$: 50
    \item Dimension of aligned encoding $d$: 50
    \item Dimension of output representation $d_{out}$: 172
    \item Number of attention heads $I$: 2
    \item Number of Transformer layers $L$: 2
    \end{itemize}
\end{itemize}

Then, we perform the grid search to find the best settings of some critical hyperparameters, where the searched ranges and related methods are shown in \tabref{tab:searched_ranges_related_methods}. It is worth noticing that DyGFormer can directly handle nodes with sequence lengths shorter than the defined length. When the sequence length exceeds the specified length, we select the most recent interactions up to the defined length. 
\begin{table}[!htbp]
\centering
\caption{Searched ranges of hyperparameters and the related methods.}
\label{tab:searched_ranges_related_methods}
\setlength{\tabcolsep}{2.0mm}
{
\begin{tabular}{c|cc}
\hline
Hyperparameters                                                             & Searched Ranges                                                                                                                        & Related Methods                                                                                      \\ \hline
Dropout Rate \cite{DBLP:journals/jmlr/SrivastavaHKSS14}                                                              & \begin{tabular}[c]{@{}c@{}}[0.0, 0.1, 0.2, 0.3, \\ 0.4, 0.5, 0.6]\end{tabular}                                                         & \begin{tabular}[c]{@{}c@{}}JODIE, DyRep, TGAT, TGN, CAWN, \\ TCL, GraphMixer, DyGFormer\end{tabular} \\
\begin{tabular}[c]{@{}c@{}}Number of \\ Sampled Neighbors\end{tabular}      & [10, 20, 30]                                                                                                                           & \begin{tabular}[c]{@{}c@{}}DyRep, TGAT, TGN, \\ TCL, GraphMixer\end{tabular}                         \\
\begin{tabular}[c]{@{}c@{}}Neighbor Sampling \\ Strategies\end{tabular}     & [uniform,recent]                                                                                                                       & \begin{tabular}[c]{@{}c@{}}DyRep, TGAT, TGN, \\ TCL, GraphMixer\end{tabular}                         \\
\begin{tabular}[c]{@{}c@{}}Number of Causal \\ Anonymous Walks\end{tabular} & [16, 32, 64, 128]                                                                                                                      & CAWN                                                                                                 \\
\begin{tabular}[c]{@{}c@{}}Memory Updating \\ Variants\end{tabular}       & \begin{tabular}[c]{@{}c@{}}[EdgeBank$_\infty$, EdgeBank$_\text{tw-ts}$, \\ EdgeBank$_\text{tw-re}$, EdgeBank$_\text{th}$]\end{tabular} & EdgeBank                                                                                             \\
\begin{tabular}[c]{@{}c@{}}Length of Input \\ Sequences \& \\ Patch Size\end{tabular} & \begin{tabular}[c]{@{}c@{}}[32 \& 1, 64 \& 2, 128 \& 4, \\ 256 \& 8, 512 \& 16, 1024 \& 32, \\ 2048 \& 64, 4096 \& 128]\end{tabular}   & DyGFormer                                                                                            \\ \hline
\end{tabular}
}
\end{table}

Finally, we show the hyperparameter settings of various methods that are determined by the grid search in \tabref{tab:dropout_configuration}, \tabref{tab:num_neighbors_configuration}, \tabref{tab:neighbors_sampling_configuration} and \tabref{tab:edgebank_configuration}.

\begin{table}[!htbp]
\centering
\caption{Configurations of the dropout rate of different methods.}
\label{tab:dropout_configuration}
\setlength{\tabcolsep}{2.0mm}
{
\begin{tabular}{c|cccccccc}
\hline
Datasets    & JODIE & DyRep & TGAT & TGN & CAWN & TCL & GraphMixer & DyGFormer \\ \hline
Wikipedia   & 0.1   & 0.1   & 0.1  & 0.1 & 0.1  & 0.1 & 0.5        & 0.1       \\
Reddit      & 0.1   & 0.1   & 0.1  & 0.1 & 0.1  & 0.1 & 0.5        & 0.2       \\
MOOC        & 0.2   & 0.0   & 0.1  & 0.2 & 0.1  & 0.1 & 0.4        & 0.1       \\
LastFM      & 0.3   & 0.0   & 0.1  & 0.3 & 0.1  & 0.1 & 0.0        & 0.1       \\
Enron       & 0.1   & 0.0   & 0.2  & 0.0 & 0.1  & 0.1 & 0.5        & 0.0       \\
Social Evo. & 0.1   & 0.1   & 0.1  & 0.0 & 0.1  & 0.0 & 0.3        & 0.1       \\
UCI         & 0.4   & 0.0   & 0.1  & 0.1 & 0.1  & 0.0 & 0.4        & 0.1       \\
Flights     & 0.1   & 0.1   & 0.1  & 0.1 & 0.1  & 0.1 & 0.2        & 0.1       \\
Can. Parl.  & 0.0   & 0.0   & 0.2  & 0.3 & 0.0  & 0.2 & 0.2        & 0.1       \\
US Legis.   & 0.2   & 0.0   & 0.1  & 0.1 & 0.1  & 0.3 & 0.4        & 0.0       \\
UN Trade    & 0.4   & 0.1   & 0.1  & 0.2 & 0.1  & 0.0 & 0.1        & 0.0       \\
UN Vote     & 0.1   & 0.1   & 0.2  & 0.1 & 0.1  & 0.0 & 0.0        & 0.2       \\
Contact     & 0.1   & 0.0   & 0.1  & 0.1 & 0.1  & 0.0 & 0.1        & 0.0       \\ \hline
\end{tabular}
}
\end{table}

\begin{table}[!htbp]
\centering
\caption{Configurations of the number of sampled neighbors, the number of causal anonymous walks, and the length of input sequences $\&$ the patch size of different methods.}
\label{tab:num_neighbors_configuration}
\begin{tabular}{c|ccccccc}
\hline
Datasets    & DyRep & TGAT & TGN & CAWN & TCL & GraphMixer & DyGFormer  \\ \hline
Wikipedia   & 10    & 20   & 10  & 32   & 20  & 30         & 32 \& 1    \\
Reddit      & 10    & 20   & 10  & 32   & 20  & 10         & 64 \& 2    \\
MOOC        & 10    & 20   & 10  & 64   & 20  & 20         & 256 \& 8   \\
LastFM      & 10    & 20   & 10  & 128  & 20  & 10         & 512 \& 16  \\
Enron       & 10    & 20   & 10  & 32   & 20  & 20         & 256 \& 8   \\
Social Evo. & 10    & 20   & 10  & 64   & 20  & 20         & 32 \& 1    \\
UCI         & 10    & 20   & 10  & 64   & 20  & 20         & 32 \& 1    \\
Flights     & 10    & 20   & 10  & 64   & 20  & 20         & 256 \& 8   \\
Can. Parl.  & 10    & 20   & 10  & 128  & 20  & 20         & 2048 \& 64 \\
US Legis.   & 10    & 20   & 10  & 32   & 20  & 20         & 256 \& 8   \\
UN Trade    & 10    & 20   & 10  & 64   & 20  & 20         & 256 \& 8   \\
UN Vote     & 10    & 20   & 10  & 64   & 20  & 20         & 128 \& 4   \\
Contact     & 10    & 20   & 10  & 64   & 20  & 20         & 32 \& 1    \\ \hline
\end{tabular}
\end{table}

\begin{table}[!htbp]
\centering
\caption{Configurations of neighbor sampling strategies of different methods.}
\label{tab:neighbors_sampling_configuration}
\begin{tabular}{c|ccccc}
\hline
Datasets    & DyRep   & TGAT    & TGN     & TCL     & GraphMixer \\ \hline
Wikipedia   & recent  & recent  & recent  & recent  & recent     \\
Reddit      & recent  & uniform & recent  & uniform & recent     \\
MOOC        & recent  & recent  & recent  & recent  & recent     \\
LastFM      & recent  & recent  & recent  & recent  & recent     \\
Enron       & recent  & recent  & recent  & recent  & recent     \\
Social Evo. & recent  & recent  & recent  & recent  & recent     \\
UCI         & recent  & recent  & recent  & recent  & recent     \\
Flights     & recent  & recent  & recent  & recent  & recent     \\
Can. Parl.  & uniform & uniform & uniform & uniform & uniform    \\
US Legis.   & recent  & recent  & recent  & uniform & recent     \\
UN Trade    & recent  & uniform & recent  & uniform & uniform    \\
UN Vote     & recent  & recent  & uniform & uniform & uniform    \\
Contact     & recent  & recent  & recent  & recent  & recent     \\ \hline
\end{tabular}
\end{table}

\begin{table}[!htbp]
\centering
\caption{Configurations of the variants of EdgeBank with three negative sampling strategies.}
\label{tab:edgebank_configuration}
\begin{tabular}{c|ccc}
\hline
Datasets    & Random                  & Historical              & Inductive               \\ \hline
Wikipedia   & EdgeBank$_\infty$       & EdgeBank$_\text{th}$    & EdgeBank$_\text{th}$    \\
Reddit      & EdgeBank$_\infty$       & EdgeBank$_\text{th}$    & EdgeBank$_\text{th}$    \\
MOOC        & EdgeBank$_\text{tw-ts}$ & EdgeBank$_\text{tw-re}$ & EdgeBank$_\text{th}$    \\
LastFM      & EdgeBank$_\text{tw-ts}$ & EdgeBank$_\text{tw-re}$ & EdgeBank$_\text{th}$    \\
Enron       & EdgeBank$_\text{tw-ts}$ & EdgeBank$_\text{tw-re}$ & EdgeBank$_\text{th}$    \\
Social Evo. & EdgeBank$_\text{th}$    & EdgeBank$_\text{th}$    & EdgeBank$_\text{th}$    \\
UCI         & EdgeBank$_\infty$       & EdgeBank$_\text{tw-ts}$ & EdgeBank$_\text{tw-re}$ \\
Flights     & EdgeBank$_\infty$       & EdgeBank$_\text{th}$    & EdgeBank$_\text{th}$    \\
Can. Parl.  & EdgeBank$_\text{tw-ts}$ & EdgeBank$_\text{tw-ts}$ & EdgeBank$_\text{th}$    \\
US Legis.   & EdgeBank$_\text{tw-ts}$ & EdgeBank$_\text{tw-ts}$ & EdgeBank$_\text{tw-ts}$ \\
UN Trade    & EdgeBank$_\text{tw-re}$ & EdgeBank$_\text{tw-re}$ & EdgeBank$_\text{th}$    \\
UN Vote     & EdgeBank$_\text{tw-re}$ & EdgeBank$_\text{tw-re}$ & EdgeBank$_\text{tw-re}$ \\
Contact     & EdgeBank$_\text{tw-re}$ & EdgeBank$_\text{tw-re}$ & EdgeBank$_\text{th}$    \\ \hline
\end{tabular}
\end{table}

\section{Detailed Experimental Results}
\subsection{Additional Results for Transductive Dynamic Link Prediction}\label{section-appendix-numerical-auc-roc-performance-transductive-dynamic-link-prediction}
We show the AUC-ROC for transductive dynamic link prediction with three negative sampling strategies in \tabref{tab:auc_roc_transductive_dynamic_link_prediction}.

\begin{table}[!htbp]
\centering
\caption{AUC-ROC for transductive dynamic link prediction with random, historical, and inductive negative sampling strategies.}
\label{tab:auc_roc_transductive_dynamic_link_prediction}
\resizebox{1.01\textwidth}{!}
{
\setlength{\tabcolsep}{0.9mm}
{
\begin{tabular}{c|c|ccccccccc}
\hline
NSS                    & Datasets    & JODIE        & DyRep        & TGAT         & TGN          & CAWN         & EdgeBank     & TCL          & GraphMixer   & DyGFormer    \\ \hline
\multirow{14}{*}{rnd}  & Wikipedia   & 96.33 $\pm$ 0.07 & 94.37 $\pm$ 0.09 & 96.67 $\pm$ 0.07 & 98.37 $\pm$ 0.07 & \underline{98.54 $\pm$ 0.04} & 90.78 $\pm$ 0.00 & 95.84 $\pm$ 0.18 & 96.92 $\pm$ 0.03 & \textbf{98.91 $\pm$ 0.02} \\
                       & Reddit      & 98.31 $\pm$ 0.05 & 98.17 $\pm$ 0.05 & 98.47 $\pm$ 0.02 & 98.60 $\pm$ 0.06 & \underline{99.01 $\pm$ 0.01} & 95.37 $\pm$ 0.00 & 97.42 $\pm$ 0.02 & 97.17 $\pm$ 0.02 & \textbf{99.15 $\pm$ 0.01} \\
                       & MOOC        & 83.81 $\pm$ 2.09 & 85.03 $\pm$ 0.58 & 87.11 $\pm$ 0.19 & \textbf{91.21 $\pm$ 1.15} & 80.38 $\pm$ 0.26 & 60.86 $\pm$ 0.00 & 83.12 $\pm$ 0.18 & 84.01 $\pm$ 0.17 & \underline{87.91 $\pm$ 0.58} \\
                       & LastFM      & 70.49 $\pm$ 1.66 & 71.16 $\pm$ 1.89 & 71.59 $\pm$ 0.18 & 78.47 $\pm$ 2.94 & \underline{85.92 $\pm$ 0.10} & 83.77 $\pm$ 0.00 & 64.06 $\pm$ 1.16 & 73.53 $\pm$ 0.12 & \textbf{93.05 $\pm$ 0.10} \\
                       & Enron       & 87.96 $\pm$ 0.52 & 84.89 $\pm$ 3.00 & 68.89 $\pm$ 1.10 & 88.32 $\pm$ 0.99 & \underline{90.45 $\pm$ 0.14} & 87.05 $\pm$ 0.00 & 75.74 $\pm$ 0.72 & 84.38 $\pm$ 0.21 & \textbf{93.33 $\pm$ 0.13} \\
                       & Social Evo. & 92.05 $\pm$ 0.46 & 90.76 $\pm$ 0.21 & 94.76 $\pm$ 0.16 & \underline{95.39 $\pm$ 0.17} & 87.34 $\pm$ 0.08 & 81.60 $\pm$ 0.00 & 94.84 $\pm$ 0.17 & 95.23 $\pm$ 0.07 & \textbf{96.30 $\pm$ 0.01} \\
                       & UCI         & 90.44 $\pm$ 0.49 & 68.77 $\pm$ 2.34 & 78.53 $\pm$ 0.74 & 92.03 $\pm$ 1.13 & \underline{93.87 $\pm$ 0.08} & 77.30 $\pm$ 0.00 & 87.82 $\pm$ 1.36 & 91.81 $\pm$ 0.67 & \textbf{94.49 $\pm$ 0.26} \\
                       & Flights     & 96.21 $\pm$ 1.42 & 95.95 $\pm$ 0.62 & 94.13 $\pm$ 0.17 & 98.22 $\pm$ 0.13 & \underline{98.45 $\pm$ 0.01} & 90.23 $\pm$ 0.00 & 91.21 $\pm$ 0.02 & 91.13 $\pm$ 0.01 & \textbf{98.93 $\pm$ 0.01} \\
                       & Can. Parl.  & 78.21 $\pm$ 0.23 & 73.35 $\pm$ 3.67 & 75.69 $\pm$ 0.78 & 76.99 $\pm$ 1.80 & 75.70 $\pm$ 3.27 & 64.14 $\pm$ 0.00 & 72.46 $\pm$ 3.23 & \underline{83.17 $\pm$ 0.53} & \textbf{97.76 $\pm$ 0.41} \\
                       & US Legis.   & \underline{82.85 $\pm$ 1.07} & 82.28 $\pm$ 0.32 & 75.84 $\pm$ 1.99 & \textbf{83.34 $\pm$ 0.43} & 77.16 $\pm$ 0.39 & 62.57 $\pm$ 0.00 & 76.27 $\pm$ 0.63 & 76.96 $\pm$ 0.79 & 77.90 $\pm$ 0.58 \\
                       & UN Trade    & \underline{69.62 $\pm$ 0.44} & 67.44 $\pm$ 0.83 & 64.01 $\pm$ 0.12 & 69.10 $\pm$ 1.67 & 68.54 $\pm$ 0.18 & 66.75 $\pm$ 0.00 & 64.72 $\pm$ 0.05 & 65.52 $\pm$ 0.51 & \textbf{70.20 $\pm$ 1.44} \\
                       & UN Vote     & \underline{68.53 $\pm$ 0.95} & 67.18 $\pm$ 1.04 & 52.83 $\pm$ 1.12 & \textbf{69.71 $\pm$ 2.65} & 53.09 $\pm$ 0.22 & 62.97 $\pm$ 0.00 & 51.88 $\pm$ 0.36 & 52.46 $\pm$ 0.27 & 57.12 $\pm$ 0.62 \\
                       & Contact     & 96.66 $\pm$ 0.89 & 96.48 $\pm$ 0.14 & 96.95 $\pm$ 0.08 & \underline{97.54 $\pm$ 0.35} & 89.99 $\pm$ 0.34 & 94.34 $\pm$ 0.00 & 94.15 $\pm$ 0.09 & 93.94 $\pm$ 0.02 & \textbf{98.53 $\pm$ 0.01} \\ \cline{2-11} 
                       & Avg. Rank   & 4.38         & 5.77         & 6.00         & \underline{2.54}         & 4.38         & 7.31         & 7.23         & 5.77         & \textbf{1.62}         \\ \hline
\multirow{14}{*}{hist} & Wikipedia   & 80.77 $\pm$ 0.73 & 77.74 $\pm$ 0.33 & 82.87 $\pm$ 0.22 & 82.74 $\pm$ 0.32 & 67.84 $\pm$ 0.64 & 77.27 $\pm$ 0.00 & \underline{85.76 $\pm$ 0.46} & \textbf{87.68 $\pm$ 0.17} & 78.80 $\pm$ 1.95 \\
                       & Reddit      & 80.52 $\pm$ 0.32 & 80.15 $\pm$ 0.18 & 79.33 $\pm$ 0.16 & \textbf{81.11 $\pm$ 0.19} & 80.27 $\pm$ 0.30 & 78.58 $\pm$ 0.00 & 76.49 $\pm$ 0.16 & 77.80 $\pm$ 0.12 & \underline{80.54 $\pm$ 0.29} \\
                       & MOOC        & 82.75 $\pm$ 0.83 & 81.06 $\pm$ 0.94 & 80.81 $\pm$ 0.67 & \textbf{88.00 $\pm$ 1.80} & 71.57 $\pm$ 1.07 & 61.90 $\pm$ 0.00 & 72.09 $\pm$ 0.56 & 76.68 $\pm$ 1.40 & \underline{87.04 $\pm$ 0.35} \\
                       & LastFM      & 75.22 $\pm$ 2.36 & 74.65 $\pm$ 1.98 & 64.27 $\pm$ 0.26 & 77.97 $\pm$ 3.04 & 67.88 $\pm$ 0.24 & \underline{78.09 $\pm$ 0.00} & 47.24 $\pm$ 3.13 & 64.21 $\pm$ 0.73 & \textbf{78.78 $\pm$ 0.35} \\
                       & Enron       & 75.39 $\pm$ 2.37 & 74.69 $\pm$ 3.55 & 61.85 $\pm$ 1.43 & \underline{77.09 $\pm$ 2.22} & 65.10 $\pm$ 0.34 & \textbf{79.59 $\pm$ 0.00} & 67.95 $\pm$ 0.88 & 75.27 $\pm$ 1.14 & 76.55 $\pm$ 0.52 \\
                       & Social Evo. & 90.06 $\pm$ 3.15 & 93.12 $\pm$ 0.34 & 93.08 $\pm$ 0.59 & \underline{94.71 $\pm$ 0.53} & 87.43 $\pm$ 0.15 & 85.81 $\pm$ 0.00 & 93.44 $\pm$ 0.68 & 94.39 $\pm$ 0.31 & \textbf{97.28 $\pm$ 0.07} \\
                       & UCI         & \textbf{78.64 $\pm$ 3.50} & 57.91 $\pm$ 3.12 & 58.89 $\pm$ 1.57 & 77.25 $\pm$ 2.68 & 57.86 $\pm$ 0.15 & 69.56 $\pm$ 0.00 & 72.25 $\pm$ 3.46 & \underline{77.54 $\pm$ 2.02} & 76.97 $\pm$ 0.24 \\
                       & Flights     & 68.97 $\pm$ 1.87 & 69.43 $\pm$ 0.90 & \underline{72.20 $\pm$ 0.16} & 68.39 $\pm$ 0.95 & 66.11 $\pm$ 0.71 & \textbf{74.64 $\pm$ 0.00} & 70.57 $\pm$ 0.18 & 70.37 $\pm$ 0.23 & 68.09 $\pm$ 0.43 \\
                       & Can. Parl.  & 62.44 $\pm$ 1.11 & 70.16 $\pm$ 1.70 & 70.86 $\pm$ 0.94 & 73.23 $\pm$ 3.08 & 72.06 $\pm$ 3.94 & 63.04 $\pm$ 0.00 & 69.95 $\pm$ 3.70 & \underline{79.03 $\pm$ 1.01} & \textbf{97.61 $\pm$ 0.40} \\
                       & US Legis.   & 67.47 $\pm$ 6.40 & \textbf{91.44 $\pm$ 1.18} & 73.47 $\pm$ 5.25 & 83.53 $\pm$ 4.53 & 78.62 $\pm$ 7.46 & 67.41 $\pm$ 0.00 & 83.97 $\pm$ 3.71 & 85.17 $\pm$ 0.70 & \underline{90.77 $\pm$ 1.96} \\
                       & UN Trade    & 68.92 $\pm$ 1.40 & 64.36 $\pm$ 1.40 & 60.37 $\pm$ 0.68 & 63.93 $\pm$ 5.41 & 63.09 $\pm$ 0.74 & \textbf{86.61 $\pm$ 0.00} & 61.43 $\pm$ 1.04 & 63.20 $\pm$ 1.54 & \underline{73.86 $\pm$ 1.13} \\
                       & UN Vote     & \underline{76.84 $\pm$ 1.01} & 74.72 $\pm$ 1.43 & 53.95 $\pm$ 3.15 & 73.40 $\pm$ 5.20 & 51.27 $\pm$ 0.33 & \textbf{89.62 $\pm$ 0.00} & 52.29 $\pm$ 2.39 & 52.61 $\pm$ 1.44 & 64.27 $\pm$ 1.78 \\
                       & Contact     & \underline{96.35 $\pm$ 0.92} & 96.00 $\pm$ 0.23 & 95.39 $\pm$ 0.43 & 93.76 $\pm$ 1.29 & 83.06 $\pm$ 0.32 & 92.17 $\pm$ 0.00 & 93.34 $\pm$ 0.19 & 93.14 $\pm$ 0.34 & \textbf{97.17 $\pm$ 0.05} \\ \cline{2-11} 
                       & Avg. Rank   & 4.38         & 4.77         & 5.85         & \underline{3.46}         & 7.38         & 5.38         & 6.08         & 4.77         & \textbf{2.92}         \\ \hline
\multirow{14}{*}{ind}  & Wikipedia   & 70.96 $\pm$ 0.78 & 67.36 $\pm$ 0.96 & 81.93 $\pm$ 0.22 & 80.97 $\pm$ 0.31 & 70.95 $\pm$ 0.95 & 81.73 $\pm$ 0.00 & \underline{82.19 $\pm$ 0.48} & \textbf{84.28 $\pm$ 0.30} & 75.09 $\pm$ 3.70 \\
                       & Reddit      & 83.51 $\pm$ 0.15 & 82.90 $\pm$ 0.31 & \underline{87.13 $\pm$ 0.20} & 84.56 $\pm$ 0.24 & \textbf{88.04 $\pm$ 0.29} & 85.93 $\pm$ 0.00 & 84.67 $\pm$ 0.29 & 82.21 $\pm$ 0.13 & 86.23 $\pm$ 0.51 \\
                       & MOOC        & 66.63 $\pm$ 2.30 & 63.26 $\pm$ 1.01 & 73.18 $\pm$ 0.33 & \underline{77.44 $\pm$ 2.86} & 70.32 $\pm$ 1.43 & 48.18 $\pm$ 0.00 & 70.36 $\pm$ 0.37 & 72.45 $\pm$ 0.72 & \textbf{80.76 $\pm$ 0.76} \\
                       & LastFM      & 61.32 $\pm$ 3.49 & 62.15 $\pm$ 2.12 & 63.99 $\pm$ 0.21 & 65.46 $\pm$ 4.27 & 67.92 $\pm$ 0.44 & \textbf{77.37 $\pm$ 0.00} & 46.93 $\pm$ 2.59 & 60.22 $\pm$ 0.32 & \underline{69.25 $\pm$ 0.36} \\
                       & Enron       & 70.92 $\pm$ 1.05 & 68.73 $\pm$ 1.34 & 60.45 $\pm$ 2.12 & 71.34 $\pm$ 2.46 & \textbf{75.17 $\pm$ 0.50} & \underline{75.00 $\pm$ 0.00} & 67.64 $\pm$ 0.86 & 71.53 $\pm$ 0.85 & 74.07 $\pm$ 0.64 \\
                       & Social Evo. & 90.01 $\pm$ 3.19 & 93.07 $\pm$ 0.38 & 92.94 $\pm$ 0.61 & \underline{95.24 $\pm$ 0.56} & 89.93 $\pm$ 0.15 & 87.88 $\pm$ 0.00 & 93.44 $\pm$ 0.72 & 94.22 $\pm$ 0.32 & \textbf{97.51 $\pm$ 0.06} \\
                       & UCI         & 64.14 $\pm$ 1.26 & 54.25 $\pm$ 2.01 & 60.80 $\pm$ 1.01 & 64.11 $\pm$ 1.04 & 58.06 $\pm$ 0.26 & 58.03 $\pm$ 0.00 & \underline{70.05 $\pm$ 1.86} & \textbf{74.59 $\pm$ 0.74} & 65.96 $\pm$ 1.18 \\
                       & Flights     & 69.99 $\pm$ 3.10 & 71.13 $\pm$ 1.55 & \underline{73.47 $\pm$ 0.18} & 71.63 $\pm$ 1.72 & 69.70 $\pm$ 0.75 & \textbf{81.10 $\pm$ 0.00} & 72.54 $\pm$ 0.19 & 72.21 $\pm$ 0.21 & 69.53 $\pm$ 1.17 \\
                       & Can. Parl.  & 52.88 $\pm$ 0.80 & 63.53 $\pm$ 0.65 & 72.47 $\pm$ 1.18 & 69.57 $\pm$ 2.81 & \underline{72.93 $\pm$ 1.78} & 61.41 $\pm$ 0.00 & 69.47 $\pm$ 2.12 & 70.52 $\pm$ 0.94 & \textbf{96.70 $\pm$ 0.59} \\
                       & US Legis.   & 59.05 $\pm$ 5.52 & \textbf{89.44 $\pm$ 0.71} & 71.62 $\pm$ 5.42 & 78.12 $\pm$ 4.46 & 76.45 $\pm$ 7.02 & 68.66 $\pm$ 0.00 & 82.54 $\pm$ 3.91 & 84.22 $\pm$ 0.91 & \underline{87.96 $\pm$ 1.80} \\
                       & UN Trade    & 66.82 $\pm$ 1.27 & 65.60 $\pm$ 1.28 & 66.13 $\pm$ 0.78 & 66.37 $\pm$ 5.39 & \underline{71.73 $\pm$ 0.74} & \textbf{74.20 $\pm$ 0.00} & 67.80 $\pm$ 1.21 & 66.53 $\pm$ 1.22 & 62.56 $\pm$ 1.51 \\
                       & UN Vote     & \textbf{73.73 $\pm$ 1.61} & 72.80 $\pm$ 2.16 & 53.04 $\pm$ 2.58 & 72.69 $\pm$ 3.72 & 52.75 $\pm$ 0.90 & \underline{72.85 $\pm$ 0.00} & 52.02 $\pm$ 1.64 & 51.89 $\pm$ 0.74 & 53.37 $\pm$ 1.26 \\
                       & Contact     & \underline{94.47 $\pm$ 1.08} & 94.23 $\pm$ 0.18 & 94.10 $\pm$ 0.41 & 91.64 $\pm$ 1.72 & 87.68 $\pm$ 0.24 & 85.87 $\pm$ 0.00 & 91.23 $\pm$ 0.19 & 90.96 $\pm$ 0.27 & \textbf{95.01 $\pm$ 0.15} \\ \cline{2-11} 
                       & Avg. Rank   & 5.92         & 6.15         & 4.85         & \underline{4.54}         & 5.15         & 5.08         & 5.00         & 4.77         & \textbf{3.54}         \\ \hline
\end{tabular}
}
}
\end{table}

\subsection{Additional Results for Inductive Dynamic Link Prediction}\label{section-appendix-numerical-performance-inductive-dynamic-link-prediction}
We present the AP and AUC-ROC for inductive dynamic link prediction with three negative sampling strategies in \tabref{tab:ap_inductive_dynamic_link_prediction} and \tabref{tab:auc_roc_inductive_dynamic_link_prediction} .

\begin{table}[!htbp]
\centering
\caption{AP for inductive dynamic link prediction with random, historical, and inductive negative sampling strategies.}
\label{tab:ap_inductive_dynamic_link_prediction}
\resizebox{1.01\textwidth}{!}
{
\setlength{\tabcolsep}{0.9mm}
{
\begin{tabular}{c|c|cccccccc}
\hline
NSS                    & Datasets    & JODIE        & DyRep        & TGAT         & TGN          & CAWN         & TCL          & GraphMixer   & DyGFormer    \\ \hline
\multirow{14}{*}{rnd}  & Wikipedia   & 94.82 $\pm$ 0.20 & 92.43 $\pm$ 0.37 & 96.22 $\pm$ 0.07 & 97.83 $\pm$ 0.04 & \underline{98.24 $\pm$ 0.03} & 96.22 $\pm$ 0.17 & 96.65 $\pm$ 0.02 & \textbf{98.59 $\pm$ 0.03} \\
                       & Reddit      & 96.50 $\pm$ 0.13 & 96.09 $\pm$ 0.11 & 97.09 $\pm$ 0.04 & 97.50 $\pm$ 0.07 & \underline{98.62 $\pm$ 0.01} & 94.09 $\pm$ 0.07 & 95.26 $\pm$ 0.02 & \textbf{98.84 $\pm$ 0.02} \\
                       & MOOC        & 79.63 $\pm$ 1.92 & 81.07 $\pm$ 0.44 & 85.50 $\pm$ 0.19 & \textbf{89.04 $\pm$ 1.17} & 81.42 $\pm$ 0.24 & 80.60 $\pm$ 0.22 & 81.41 $\pm$ 0.21 & \underline{86.96 $\pm$ 0.43} \\
                       & LastFM      & 81.61 $\pm$ 3.82 & 83.02 $\pm$ 1.48 & 78.63 $\pm$ 0.31 & 81.45 $\pm$ 4.29 & \underline{89.42 $\pm$ 0.07} & 73.53 $\pm$ 1.66 & 82.11 $\pm$ 0.42 & \textbf{94.23 $\pm$ 0.09} \\
                       & Enron       & 80.72 $\pm$ 1.39 & 74.55 $\pm$ 3.95 & 67.05 $\pm$ 1.51 & 77.94 $\pm$ 1.02 & \underline{86.35 $\pm$ 0.51} & 76.14 $\pm$ 0.79 & 75.88 $\pm$ 0.48 & \textbf{89.76 $\pm$ 0.34} \\
                       & Social Evo. & \underline{91.96 $\pm$ 0.48} & 90.04 $\pm$ 0.47 & 91.41 $\pm$ 0.16 & 90.77 $\pm$ 0.86 & 79.94 $\pm$ 0.18 & 91.55 $\pm$ 0.09 & 91.86 $\pm$ 0.06 & \textbf{93.14 $\pm$ 0.04} \\
                       & UCI         & 79.86 $\pm$ 1.48 & 57.48 $\pm$ 1.87 & 79.54 $\pm$ 0.48 & 88.12 $\pm$ 2.05 & \underline{92.73 $\pm$ 0.06} & 87.36 $\pm$ 2.03 & 91.19 $\pm$ 0.42 & \textbf{94.54 $\pm$ 0.12} \\
                       & Flights     & 94.74 $\pm$ 0.37 & 92.88 $\pm$ 0.73 & 88.73 $\pm$ 0.33 & 95.03 $\pm$ 0.60 & \underline{97.06 $\pm$ 0.02} & 83.41 $\pm$ 0.07 & 83.03 $\pm$ 0.05 & \textbf{97.79 $\pm$ 0.02} \\
                       & Can. Parl.  & 53.92 $\pm$ 0.94 & 54.02 $\pm$ 0.76 & 55.18 $\pm$ 0.79 & 54.10 $\pm$ 0.93 & 55.80 $\pm$ 0.69 & 54.30 $\pm$ 0.66 & \underline{55.91 $\pm$ 0.82} & \textbf{87.74 $\pm$ 0.71} \\
                       & US Legis.   & 54.93 $\pm$ 2.29 & \underline{57.28 $\pm$ 0.71} & 51.00 $\pm$ 3.11 & \textbf{58.63 $\pm$ 0.37} & 53.17 $\pm$ 1.20 & 52.59 $\pm$ 0.97 & 50.71 $\pm$ 0.76 & 54.28 $\pm$ 2.87 \\
                       & UN Trade    & 59.65 $\pm$ 0.77 & 57.02 $\pm$ 0.69 & 61.03 $\pm$ 0.18 & 58.31 $\pm$ 3.15 & \textbf{65.24 $\pm$ 0.21} & 62.21 $\pm$ 0.12 & 62.17 $\pm$ 0.31 & \underline{64.55 $\pm$ 0.62} \\
                       & UN Vote     & \underline{56.64 $\pm$ 0.96} & 54.62 $\pm$ 2.22 & 52.24 $\pm$ 1.46 & \textbf{58.85 $\pm$ 2.51} & 49.94 $\pm$ 0.45 & 51.60 $\pm$ 0.97 & 50.68 $\pm$ 0.44 & 55.93 $\pm$ 0.39 \\
                       & Contact     & 94.34 $\pm$ 1.45 & 92.18 $\pm$ 0.41 & \underline{95.87 $\pm$ 0.11} & 93.82 $\pm$ 0.99 & 89.55 $\pm$ 0.30 & 91.11 $\pm$ 0.12 & 90.59 $\pm$ 0.05 & \textbf{98.03 $\pm$ 0.02} \\ \cline{2-10} 
                       & Avg. Rank   & 4.69         & 5.77         & 5.23         & \underline{3.77}         & \underline{3.77}         & 5.77         & 5.23         & \textbf{1.54}         \\ \hline
\multirow{14}{*}{hist} & Wikipedia   & 68.69 $\pm$ 0.39 & 62.18 $\pm$ 1.27 & \underline{84.17 $\pm$ 0.22} & 81.76 $\pm$ 0.32 & 67.27 $\pm$ 1.63 & 82.20 $\pm$ 2.18 & \textbf{87.60 $\pm$ 0.30} & 71.42 $\pm$ 4.43 \\
                       & Reddit      & 62.34 $\pm$ 0.54 & 61.60 $\pm$ 0.72 & 63.47 $\pm$ 0.36 & \underline{64.85 $\pm$ 0.85} & 63.67 $\pm$ 0.41 & 60.83 $\pm$ 0.25 & 64.50 $\pm$ 0.26 & \textbf{65.37 $\pm$ 0.60} \\
                       & MOOC        & 63.22 $\pm$ 1.55 & 62.93 $\pm$ 1.24 & 76.73 $\pm$ 0.29 & \underline{77.07 $\pm$ 3.41} & 74.68 $\pm$ 0.68 & 74.27 $\pm$ 0.53 & 74.00 $\pm$ 0.97 & \textbf{80.82 $\pm$ 0.30} \\
                       & LastFM      & 70.39 $\pm$ 4.31 & 71.45 $\pm$ 1.76 & 76.27 $\pm$ 0.25 & 66.65 $\pm$ 6.11 & 71.33 $\pm$ 0.47 & 65.78 $\pm$ 0.65 & \textbf{76.42 $\pm$ 0.22} & \underline{76.35 $\pm$ 0.52} \\
                       & Enron       & 65.86 $\pm$ 3.71 & 62.08 $\pm$ 2.27 & 61.40 $\pm$ 1.31 & 62.91 $\pm$ 1.16 & 60.70 $\pm$ 0.36 & \underline{67.11 $\pm$ 0.62} & \textbf{72.37 $\pm$ 1.37} & 67.07 $\pm$ 0.62 \\
                       & Social Evo. & 88.51 $\pm$ 0.87 & 88.72 $\pm$ 1.10 & 93.97 $\pm$ 0.54 & 90.66 $\pm$ 1.62 & 79.83 $\pm$ 0.38 & \underline{94.10 $\pm$ 0.31} & 94.01 $\pm$ 0.47 & \textbf{96.82 $\pm$ 0.16} \\
                       & UCI         & 63.11 $\pm$ 2.27 & 52.47 $\pm$ 2.06 & 70.52 $\pm$ 0.93 & 70.78 $\pm$ 0.78 & 64.54 $\pm$ 0.47 & \underline{76.71 $\pm$ 1.00} & \textbf{81.66 $\pm$ 0.49} & 72.13 $\pm$ 1.87 \\
                       & Flights     & 61.01 $\pm$ 1.65 & 62.83 $\pm$ 1.31 & \underline{64.72 $\pm$ 0.36} & 59.31 $\pm$ 1.43 & 56.82 $\pm$ 0.57 & 64.50 $\pm$ 0.25 & \textbf{65.28 $\pm$ 0.24} & 57.11 $\pm$ 0.21 \\
                       & Can. Parl.  & 52.60 $\pm$ 0.88 & 52.28 $\pm$ 0.31 & 56.72 $\pm$ 0.47 & 54.42 $\pm$ 0.77 & \underline{57.14 $\pm$ 0.07} & 55.71 $\pm$ 0.74 & 55.84 $\pm$ 0.73 & \textbf{87.40 $\pm$ 0.85} \\
                       & US Legis.   & 52.94 $\pm$ 2.11 & \textbf{62.10 $\pm$ 1.41} & 51.83 $\pm$ 3.95 & \underline{61.18 $\pm$ 1.10} & 55.56 $\pm$ 1.71 & 53.87 $\pm$ 1.41 & 52.03 $\pm$ 1.02 & 56.31 $\pm$ 3.46 \\
                       & UN Trade    & 55.46 $\pm$ 1.19 & \underline{55.49 $\pm$ 0.84} & 55.28 $\pm$ 0.71 & 52.80 $\pm$ 3.19 & 55.00 $\pm$ 0.38 & \textbf{55.76 $\pm$ 1.03} & 54.94 $\pm$ 0.97 & 53.20 $\pm$ 1.07 \\
                       & UN Vote     & \underline{61.04 $\pm$ 1.30} & 60.22 $\pm$ 1.78 & 53.05 $\pm$ 3.10 & \textbf{63.74 $\pm$ 3.00} & 47.98 $\pm$ 0.84 & 54.19 $\pm$ 2.17 & 48.09 $\pm$ 0.43 & 52.63 $\pm$ 1.26 \\
                       & Contact     & 90.42 $\pm$ 2.34 & 89.22 $\pm$ 0.66 & \textbf{94.15 $\pm$ 0.45} & 88.13 $\pm$ 1.50 & 74.20 $\pm$ 0.80 & 90.44 $\pm$ 0.17 & 89.91 $\pm$ 0.36 & \underline{93.56 $\pm$ 0.52} \\ \cline{2-10} 
                       & Avg. Rank   & 5.38         & 5.46         & 4.00         & 4.54         & 5.92         & 3.92         & \underline{3.54}         & \textbf{3.23}         \\ \hline
\multirow{14}{*}{ind}  & Wikipedia   & 68.70 $\pm$ 0.39 & 62.19 $\pm$ 1.28 & \underline{84.17 $\pm$ 0.22} & 81.77 $\pm$ 0.32 & 67.24 $\pm$ 1.63 & 82.20 $\pm$ 2.18 & \textbf{87.60 $\pm$ 0.29} & 71.42 $\pm$ 4.43 \\
                       & Reddit      & 62.32 $\pm$ 0.54 & 61.58 $\pm$ 0.72 & 63.40 $\pm$ 0.36 & \underline{64.84 $\pm$ 0.84} & 63.65 $\pm$ 0.41 & 60.81 $\pm$ 0.26 & 64.49 $\pm$ 0.25 & \textbf{65.35 $\pm$ 0.60} \\
                       & MOOC        & 63.22 $\pm$ 1.55 & 62.92 $\pm$ 1.24 & 76.72 $\pm$ 0.30 & \underline{77.07 $\pm$ 3.40} & 74.69 $\pm$ 0.68 & 74.28 $\pm$ 0.53 & 73.99 $\pm$ 0.97 & \textbf{80.82 $\pm$ 0.30} \\
                       & LastFM      & 70.39 $\pm$ 4.31 & 71.45 $\pm$ 1.75 & 76.28 $\pm$ 0.25 & 69.46 $\pm$ 4.65 & 71.33 $\pm$ 0.47 & 65.78 $\pm$ 0.65 & \textbf{76.42 $\pm$ 0.22} & \underline{76.35 $\pm$ 0.52} \\
                       & Enron       & 65.86 $\pm$ 3.71 & 62.08 $\pm$ 2.27 & 61.40 $\pm$ 1.30 & 62.90 $\pm$ 1.16 & 60.72 $\pm$ 0.36 & \underline{67.11 $\pm$ 0.62} & \textbf{72.37 $\pm$ 1.38} & 67.07 $\pm$ 0.62 \\
                       & Social Evo. & 88.51 $\pm$ 0.87 & 88.72 $\pm$ 1.10 & 93.97 $\pm$ 0.54 & 90.65 $\pm$ 1.62 & 79.83 $\pm$ 0.39 & \underline{94.10 $\pm$ 0.32} & 94.01 $\pm$ 0.47 & \textbf{96.82 $\pm$ 0.17} \\
                       & UCI         & 63.16 $\pm$ 2.27 & 52.47 $\pm$ 2.09 & 70.49 $\pm$ 0.93 & 70.73 $\pm$ 0.79 & 64.54 $\pm$ 0.47 & \underline{76.65 $\pm$ 0.99} & \textbf{81.64 $\pm$ 0.49} & 72.13 $\pm$ 1.86 \\
                       & Flights     & 61.01 $\pm$ 1.66 & 62.83 $\pm$ 1.31 & \underline{64.72 $\pm$ 0.37} & 59.32 $\pm$ 1.45 & 56.82 $\pm$ 0.56 & 64.50 $\pm$ 0.25 & \textbf{65.29 $\pm$ 0.24} & 57.11 $\pm$ 0.20 \\
                       & Can. Parl.  & 52.58 $\pm$ 0.86 & 52.24 $\pm$ 0.28 & 56.46 $\pm$ 0.50 & 54.18 $\pm$ 0.73 & \underline{57.06 $\pm$ 0.08} & 55.46 $\pm$ 0.69 & 55.76 $\pm$ 0.65 & \textbf{87.22 $\pm$ 0.82} \\
                       & US Legis.   & 52.94 $\pm$ 2.11 & \textbf{62.10 $\pm$ 1.41} & 51.83 $\pm$ 3.95 & \underline{61.18 $\pm$ 1.10} & 55.56 $\pm$ 1.71 & 53.87 $\pm$ 1.41 & 52.03 $\pm$ 1.02 & 56.31 $\pm$ 3.46 \\
                       & UN Trade    & 55.43 $\pm$ 1.20 & 55.42 $\pm$ 0.87 & \underline{55.58 $\pm$ 0.68} & 52.80 $\pm$ 3.24 & 54.97 $\pm$ 0.38 & \textbf{55.66 $\pm$ 0.98} & 54.88 $\pm$ 1.01 & 52.56 $\pm$ 1.70 \\
                       & UN Vote     & \underline{61.17 $\pm$ 1.33} & 60.29 $\pm$ 1.79 & 53.08 $\pm$ 3.10 & \textbf{63.71 $\pm$ 2.97} & 48.01 $\pm$ 0.82 & 54.13 $\pm$ 2.16 & 48.10 $\pm$ 0.40 & 52.61 $\pm$ 1.25 \\
                       & Contact     & 90.43 $\pm$ 2.33 & 89.22 $\pm$ 0.65 & \textbf{94.14 $\pm$ 0.45} & 88.12 $\pm$ 1.50 & 74.19 $\pm$ 0.81 & 90.43 $\pm$ 0.17 & 89.91 $\pm$ 0.36 & \underline{93.55 $\pm$ 0.52} \\ \cline{2-10} 
                       & Avg. Rank   & 5.31         & 5.54         & 3.85         & 4.38         & 5.85         & 3.92         & \underline{3.46}         & \textbf{3.31}         \\ \hline
\end{tabular}
}
}
\end{table}

\begin{table}[!htbp]
\centering
\caption{AUC-ROC for inductive dynamic link prediction with random, historical, and inductive negative sampling strategies.}
\label{tab:auc_roc_inductive_dynamic_link_prediction}
\resizebox{1.01\textwidth}{!}
{
\setlength{\tabcolsep}{0.9mm}
{
\begin{tabular}{c|c|cccccccc}
\hline
NSS                    & Datasets    & JODIE        & DyRep        & TGAT         & TGN          & CAWN         & TCL          & GraphMixer   & DyGFormer    \\ \hline
\multirow{14}{*}{rnd}  & Wikipedia   & 94.33 $\pm$ 0.27 & 91.49 $\pm$ 0.45 & 95.90 $\pm$ 0.09 & 97.72 $\pm$ 0.03 & \underline{98.03 $\pm$ 0.04} & 95.57 $\pm$ 0.20 & 96.30 $\pm$ 0.04 & \textbf{98.48 $\pm$ 0.03} \\
                       & Reddit      & 96.52 $\pm$ 0.13 & 96.05 $\pm$ 0.12 & 96.98 $\pm$ 0.04 & 97.39 $\pm$ 0.07 & \underline{98.42 $\pm$ 0.02} & 93.80 $\pm$ 0.07 & 94.97 $\pm$ 0.05 & \textbf{98.71 $\pm$ 0.01} \\
                       & MOOC        & 83.16 $\pm$ 1.30 & 84.03 $\pm$ 0.49 & 86.84 $\pm$ 0.17 & \textbf{91.24 $\pm$ 0.99} & 81.86 $\pm$ 0.25 & 81.43 $\pm$ 0.19 & 82.77 $\pm$ 0.24 & \underline{87.62 $\pm$ 0.51} \\
                       & LastFM      & 81.13 $\pm$ 3.39 & 82.24 $\pm$ 1.51 & 76.99 $\pm$ 0.29 & 82.61 $\pm$ 3.15 & \underline{87.82 $\pm$ 0.12} & 70.84 $\pm$ 0.85 & 80.37 $\pm$ 0.18 & \textbf{94.08 $\pm$ 0.08} \\
                       & Enron       & 81.96 $\pm$ 1.34 & 76.34 $\pm$ 4.20 & 64.63 $\pm$ 1.74 & 78.83 $\pm$ 1.11 & \underline{87.02 $\pm$ 0.50} & 72.33 $\pm$ 0.99 & 76.51 $\pm$ 0.71 & \textbf{90.69 $\pm$ 0.26} \\
                       & Social Evo. & 93.70 $\pm$ 0.29 & 91.18 $\pm$ 0.49 & 93.41 $\pm$ 0.19 & 93.43 $\pm$ 0.59 & 84.73 $\pm$ 0.27 & 93.71 $\pm$ 0.18 & \underline{94.09 $\pm$ 0.07} & \textbf{95.29 $\pm$ 0.03} \\
                       & UCI         & 78.80 $\pm$ 0.94 & 58.08 $\pm$ 1.81 & 77.64 $\pm$ 0.38 & 86.68 $\pm$ 2.29 & \underline{90.40 $\pm$ 0.11} & 84.49 $\pm$ 1.82 & 89.30 $\pm$ 0.57 & \textbf{92.63 $\pm$ 0.13} \\
                       & Flights     & 95.21 $\pm$ 0.32 & 93.56 $\pm$ 0.70 & 88.64 $\pm$ 0.35 & 95.92 $\pm$ 0.43 & \underline{96.86 $\pm$ 0.02} & 82.48 $\pm$ 0.01 & 82.27 $\pm$ 0.06 & \textbf{97.80 $\pm$ 0.02} \\
                       & Can. Parl.  & 53.81 $\pm$ 1.14 & 55.27 $\pm$ 0.49 & 56.51 $\pm$ 0.75 & 55.86 $\pm$ 0.75 & \underline{58.83 $\pm$ 1.13} & 55.83 $\pm$ 1.07 & 58.32 $\pm$ 1.08 & \textbf{89.33 $\pm$ 0.48} \\
                       & US Legis.   & 58.12 $\pm$ 2.35 & \underline{61.07 $\pm$ 0.56} & 48.27 $\pm$ 3.50 & \textbf{62.38 $\pm$ 0.48} & 51.49 $\pm$ 1.13 & 50.43 $\pm$ 1.48 & 47.20 $\pm$ 0.89 & 53.21 $\pm$ 3.04 \\
                       & UN Trade    & 62.28 $\pm$ 0.50 & 58.82 $\pm$ 0.98 & 62.72 $\pm$ 0.12 & 59.99 $\pm$ 3.50 & \underline{67.05 $\pm$ 0.21} & 63.76 $\pm$ 0.07 & 63.48 $\pm$ 0.37 & \textbf{67.25 $\pm$ 1.05} \\
                       & UN Vote     & \underline{58.13 $\pm$ 1.43} & 55.13 $\pm$ 3.46 & 51.83 $\pm$ 1.35 & \textbf{61.23 $\pm$ 2.71} & 48.34 $\pm$ 0.76 & 50.51 $\pm$ 1.05 & 50.04 $\pm$ 0.86 & 56.73 $\pm$ 0.69 \\
                       & Contact     & 95.37 $\pm$ 0.92 & 91.89 $\pm$ 0.38 & \underline{96.53 $\pm$ 0.10} & 94.84 $\pm$ 0.75 & 89.07 $\pm$ 0.34 & 93.05 $\pm$ 0.09 & 92.83 $\pm$ 0.05 & \textbf{98.30 $\pm$ 0.02} \\ \cline{2-10} 
                       & Avg. Rank   & 4.69         & 5.85         & 5.31         & \underline{3.38}         & 4.00         & 6.00         & 5.31         & \textbf{1.46}         \\ \hline
\multirow{14}{*}{hist} & Wikipedia   & 61.86 $\pm$ 0.53 & 57.54 $\pm$ 1.09 & 78.38 $\pm$ 0.20 & 75.75 $\pm$ 0.29 & 62.04 $\pm$ 0.65 & \underline{79.79 $\pm$ 0.96} & \textbf{82.87 $\pm$ 0.21} & 68.33 $\pm$ 2.82 \\
                       & Reddit      & 61.69 $\pm$ 0.39 & 60.45 $\pm$ 0.37 & 64.43 $\pm$ 0.27 & 64.55 $\pm$ 0.50 & \textbf{64.94 $\pm$ 0.21} & 61.43 $\pm$ 0.26 & 64.27 $\pm$ 0.13 & \underline{64.81 $\pm$ 0.25} \\
                       & MOOC        & 64.48 $\pm$ 1.64 & 64.23 $\pm$ 1.29 & 74.08 $\pm$ 0.27 & \underline{77.69 $\pm$ 3.55} & 71.68 $\pm$ 0.94 & 69.82 $\pm$ 0.32 & 72.53 $\pm$ 0.84 & \textbf{80.77 $\pm$ 0.63} \\
                       & LastFM      & 68.44 $\pm$ 3.26 & 68.79 $\pm$ 1.08 & 69.89 $\pm$ 0.28 & 66.99 $\pm$ 5.62 & 67.69 $\pm$ 0.24 & 55.88 $\pm$ 1.85 & \underline{70.07 $\pm$ 0.20} & \textbf{70.73 $\pm$ 0.37} \\
                       & Enron       & 65.32 $\pm$ 3.57 & 61.50 $\pm$ 2.50 & 57.84 $\pm$ 2.18 & 62.68 $\pm$ 1.09 & 62.25 $\pm$ 0.40 & 64.06 $\pm$ 1.02 & \textbf{68.20 $\pm$ 1.62} & \underline{65.78 $\pm$ 0.42} \\
                       & Social Evo. & 88.53 $\pm$ 0.55 & 87.93 $\pm$ 1.05 & 91.87 $\pm$ 0.72 & 92.10 $\pm$ 1.22 & 83.54 $\pm$ 0.24 & 93.28 $\pm$ 0.60 & \underline{93.62 $\pm$ 0.35} & \textbf{96.91 $\pm$ 0.09} \\
                       & UCI         & 60.24 $\pm$ 1.94 & 51.25 $\pm$ 2.37 & 62.32 $\pm$ 1.18 & 62.69 $\pm$ 0.90 & 56.39 $\pm$ 0.10 & \underline{70.46 $\pm$ 1.94} & \textbf{75.98 $\pm$ 0.84} & 65.55 $\pm$ 1.01 \\
                       & Flights     & 60.72 $\pm$ 1.29 & 61.99 $\pm$ 1.39 & \underline{63.38 $\pm$ 0.26} & 59.66 $\pm$ 1.04 & 56.58 $\pm$ 0.44 & \textbf{63.48 $\pm$ 0.23} & 63.30 $\pm$ 0.19 & 56.05 $\pm$ 0.21 \\
                       & Can. Parl.  & 51.62 $\pm$ 1.00 & 52.38 $\pm$ 0.46 & 58.30 $\pm$ 0.61 & 55.64 $\pm$ 0.54 & \underline{60.11 $\pm$ 0.48} & 57.30 $\pm$ 1.03 & 56.68 $\pm$ 1.20 & \textbf{88.68 $\pm$ 0.74} \\
                       & US Legis.   & 58.12 $\pm$ 2.94 & \textbf{67.94 $\pm$ 0.98} & 49.99 $\pm$ 4.88 & \underline{64.87 $\pm$ 1.65} & 54.41 $\pm$ 1.31 & 52.12 $\pm$ 2.13 & 49.28 $\pm$ 0.86 & 56.57 $\pm$ 3.22 \\
                       & UN Trade    & 58.73 $\pm$ 1.19 & 57.90 $\pm$ 1.33 & 59.74 $\pm$ 0.59 & 55.61 $\pm$ 3.54 & \underline{60.95 $\pm$ 0.80} & \textbf{61.12 $\pm$ 0.97} & 59.88 $\pm$ 1.17 & 58.46 $\pm$ 1.65 \\
                       & UN Vote     & \underline{65.16 $\pm$ 1.28} & 63.98 $\pm$ 2.12 & 51.73 $\pm$ 4.12 & \textbf{68.59 $\pm$ 3.11} & 48.01 $\pm$ 1.77 & 54.66 $\pm$ 2.11 & 45.49 $\pm$ 0.42 & 53.85 $\pm$ 2.02 \\
                       & Contact     & 90.80 $\pm$ 1.18 & 88.88 $\pm$ 0.68 & \underline{93.76 $\pm$ 0.41} & 88.84 $\pm$ 1.39 & 74.79 $\pm$ 0.37 & 90.37 $\pm$ 0.16 & 90.04 $\pm$ 0.29 & \textbf{94.14 $\pm$ 0.26} \\ \cline{2-10} 
                       & Avg. Rank   & 5.08         & 6.00         & 4.23         & 4.54         & 5.38         & 4.00         & \underline{3.69}         & \textbf{3.08}         \\ \hline
\multirow{14}{*}{ind}  & Wikipedia   & 61.87 $\pm$ 0.53 & 57.54 $\pm$ 1.09 & 78.38 $\pm$ 0.20 & 75.76 $\pm$ 0.29 & 62.02 $\pm$ 0.65 & \underline{79.79 $\pm$ 0.96} & \textbf{82.88 $\pm$ 0.21} & 68.33 $\pm$ 2.82 \\
                       & Reddit      & 61.69 $\pm$ 0.39 & 60.44 $\pm$ 0.37 & 64.39 $\pm$ 0.27 & 64.55 $\pm$ 0.50 & \textbf{64.91 $\pm$ 0.21} & 61.36 $\pm$ 0.26 & 64.27 $\pm$ 0.13 & \underline{64.80 $\pm$ 0.25} \\
                       & MOOC        & 64.48 $\pm$ 1.64 & 64.22 $\pm$ 1.29 & 74.07 $\pm$ 0.27 & \underline{77.68 $\pm$ 3.55} & 71.69 $\pm$ 0.94 & 69.83 $\pm$ 0.32 & 72.52 $\pm$ 0.84 & \textbf{80.77 $\pm$ 0.63} \\
                       & LastFM      & 68.44 $\pm$ 3.26 & 68.79 $\pm$ 1.08 & 69.89 $\pm$ 0.28 & 66.99 $\pm$ 5.61 & 67.68 $\pm$ 0.24 & 55.88 $\pm$ 1.85 & \underline{70.07 $\pm$ 0.20} & \textbf{70.73 $\pm$ 0.37} \\
                       & Enron       & 65.32 $\pm$ 3.57 & 61.50 $\pm$ 2.50 & 57.83 $\pm$ 2.18 & 62.68 $\pm$ 1.09 & 62.27 $\pm$ 0.40 & 64.05 $\pm$ 1.02 & \textbf{68.19 $\pm$ 1.63} & \underline{65.79 $\pm$ 0.42} \\
                       & Social Evo. & 88.53 $\pm$ 0.55 & 87.93 $\pm$ 1.05 & 91.88 $\pm$ 0.72 & 92.10 $\pm$ 1.22 & 83.54 $\pm$ 0.24 & 93.28 $\pm$ 0.60 & \underline{93.62 $\pm$ 0.35} & \textbf{96.91 $\pm$ 0.09} \\
                       & UCI         & 60.27 $\pm$ 1.94 & 51.26 $\pm$ 2.40 & 62.29 $\pm$ 1.17 & 62.66 $\pm$ 0.91 & 56.39 $\pm$ 0.11 & \underline{70.42 $\pm$ 1.93} & \textbf{75.97 $\pm$ 0.85} & 65.58 $\pm$ 1.00 \\
                       & Flights     & 60.72 $\pm$ 1.29 & 61.99 $\pm$ 1.39 & \underline{63.40 $\pm$ 0.26} & 59.66 $\pm$ 1.05 & 56.58 $\pm$ 0.44 & \textbf{63.49 $\pm$ 0.23} & 63.32 $\pm$ 0.19 & 56.05 $\pm$ 0.22 \\
                       & Can. Parl.  & 51.61 $\pm$ 0.98 & 52.35 $\pm$ 0.52 & 58.15 $\pm$ 0.62 & 55.43 $\pm$ 0.42 & \underline{60.01 $\pm$ 0.47} & 56.88 $\pm$ 0.93 & 56.63 $\pm$ 1.09 & \textbf{88.51 $\pm$ 0.73} \\
                       & US Legis.   & 58.12 $\pm$ 2.94 & \textbf{67.94 $\pm$ 0.98} & 49.99 $\pm$ 4.88 & \underline{64.87 $\pm$ 1.65} & 54.41 $\pm$ 1.31 & 52.12 $\pm$ 2.13 & 49.28 $\pm$ 0.86 & 56.57 $\pm$ 3.22 \\
                       & UN Trade    & 58.71 $\pm$ 1.20 & 57.87 $\pm$ 1.36 & 59.98 $\pm$ 0.59 & 55.62 $\pm$ 3.59 & \underline{60.88 $\pm$ 0.79} & \textbf{61.01 $\pm$ 0.93} & 59.71 $\pm$ 1.17 & 57.28 $\pm$ 3.06 \\
                       & UN Vote     & \underline{65.29 $\pm$ 1.30} & 64.10 $\pm$ 2.10 & 51.78 $\pm$ 4.14 & \textbf{68.58 $\pm$ 3.08} & 48.04 $\pm$ 1.76 & 54.65 $\pm$ 2.20 & 45.57 $\pm$ 0.41 & 53.87 $\pm$ 2.01 \\
                       & Contact     & 90.80 $\pm$ 1.18 & 88.87 $\pm$ 0.67 & \underline{93.76 $\pm$ 0.40} & 88.85 $\pm$ 1.39 & 74.79 $\pm$ 0.38 & 90.37 $\pm$ 0.16 & 90.04 $\pm$ 0.29 & \textbf{94.14 $\pm$ 0.26} \\ \cline{2-10} 
                       & Avg. Rank   & 5.08         & 5.92         & 4.15         & 4.54         & 5.38         & 4.00         & \underline{3.77}         & \textbf{3.15}         \\ \hline
\end{tabular}
}
}
\end{table}

\subsection{Results on Dynamic Node Classification}\label{section-appendix-node-clasification}
\tabref{tab:auc_roc_dynamic_node_classification} shows the results of various methods for dynamic node classification on Wikipedia and Reddit (the only two datasets with dynamic labels).


\begin{table}[!htbp]
\centering
\caption{AUC-ROC for dynamic node classification.}
\label{tab:auc_roc_dynamic_node_classification}
\resizebox{0.6\textwidth}{!}
{
\setlength{\tabcolsep}{1.5mm}
{
\begin{tabular}{c|cc|c}
\hline
Methods    & Wikipedia                & Reddit                   & Avg. Rank \\ \hline
JODIE      & \textbf{88.99 $\pm$ 1.05} & 60.37 $\pm$ 2.58          & 4.50                                                \\
DyRep      & 86.39 $\pm$ 0.98          & 63.72 $\pm$ 1.32          & 5.00                                                \\
TGAT       & 84.09 $\pm$ 1.27          & \textbf{70.04 $\pm$ 1.09} & \underline{4.00}                                                \\
TGN        & 86.38 $\pm$ 2.34          & 63.27 $\pm$ 0.90          & 6.00                                                \\
CAWN       & 84.88 $\pm$ 1.33          & 66.34 $\pm$ 1.78          & 5.00                                                \\
TCL        & 77.83 $\pm$ 2.13          & {\underline{68.87 $\pm$ 2.15}}    & 5.00                                                \\
GraphMixer & 86.80 $\pm$ 0.79          & 64.22 $\pm$ 3.32          & \underline{4.00}                                                \\
DyGFormer  & {\underline{87.44 $\pm$ 1.08}}    & 68.00 $\pm$ 1.74          & \textbf{2.50}                                                \\ \hline
\end{tabular}
}
}
\end{table}

\subsection{Complete Results of TCL and GraphMixer with Neighbor Co-occurrence Encoding}\label{section-appendix-node-cooccurrence-generalizability}
We show the complete results of TCL and GraphMixer with our neighbor co-occurrence encoding scheme in \tabref{tab:complete_generalizability_neighbor_co_occurrence_encoding}.

\begin{table}[!htbp]
\centering
\caption{AP for different methods when equipped with the neighbor co-occurrence encoding.}
\label{tab:complete_generalizability_neighbor_co_occurrence_encoding}
\resizebox{0.95\textwidth}{!}
{
\setlength{\tabcolsep}{1.0mm}
{
\begin{tabular}{c|ccccccc}
\hline
\multirow{2}{*}{Datasets} & \multicolumn{3}{c|}{TCL}                               & \multicolumn{3}{c|}{GraphMixer}                        & \multirow{2}{*}{DyGFormer} \\ \cline{2-7}
                          & Original & w/ NCoE & \multicolumn{1}{c|}{Improv.} & Original & w/ NCoE & \multicolumn{1}{c|}{Improv.} &                            \\ \hline
Wikipedia                 & 96.47   & \textbf{99.09}  & \multicolumn{1}{c|}{2.72\%}       & 97.25   & 97.90  & \multicolumn{1}{c|}{0.67\%}       & \underline{99.03}                     \\
Reddit                    & 97.53   & \underline{99.04}  & \multicolumn{1}{c|}{1.55\%}       & 97.31   & 97.63  & \multicolumn{1}{c|}{0.33\%}       & \textbf{99.22}                     \\
MOOC                      & 82.38   & \underline{86.92}  & \multicolumn{1}{c|}{5.51\%}       & 82.78   & 83.58  & \multicolumn{1}{c|}{0.97\%}       & \textbf{87.52}                     \\
LastFM                    & 67.27   & \underline{84.02}  & \multicolumn{1}{c|}{24.90\%}      & 75.61   & 76.48  & \multicolumn{1}{c|}{1.15\%}       & \textbf{93.00}                     \\
Enron                     & 79.70   & \underline{90.18}  & \multicolumn{1}{c|}{13.15\%}      & 82.25   & 88.83  & \multicolumn{1}{c|}{8.00\%}       & \textbf{92.47}                     \\
Social Evo.               & 93.13   & 94.06  & \multicolumn{1}{c|}{1.00\%}       & 93.37   & \underline{94.37}  & \multicolumn{1}{c|}{1.07\%}       & \textbf{94.73}                     \\
UCI                       & 89.57   & \underline{94.69}  & \multicolumn{1}{c|}{5.72\%}       & 93.25   & 93.48  & \multicolumn{1}{c|}{0.25\%}       & \textbf{95.79}                     \\
Flights                   & 91.23   & \underline{97.71}  & \multicolumn{1}{c|}{7.10\%}       & 90.99   & 96.90  & \multicolumn{1}{c|}{6.50\%}       & \textbf{98.91}                     \\
Can. Parl.                & 68.67   & 69.34  & \multicolumn{1}{c|}{0.98\%}       & \underline{77.04}   & 76.38  & \multicolumn{1}{c|}{-0.86\%}      & \textbf{97.36}                     \\
US Legis.                 & 69.59   & 69.47  & \multicolumn{1}{c|}{-0.17\%}      & \underline{70.74}   & 70.26  & \multicolumn{1}{c|}{-0.68\%}      & \textbf{71.11}                     \\
UN Trade                  & 62.21   & \underline{63.46}  & \multicolumn{1}{c|}{2.01\%}       & 62.61   & 62.77  & \multicolumn{1}{c|}{0.26\%}       & \textbf{66.46}                    \\
UN Vote                   & 51.90   & 51.52  & \multicolumn{1}{c|}{-0.73\%}      & 52.11   & \underline{52.13}  & \multicolumn{1}{c|}{0.04\%}       & \textbf{55.55}                     \\
Contact                   & 92.44   & \underline{97.98}  & \multicolumn{1}{c|}{5.99\%}       & 91.92   & 97.94  & \multicolumn{1}{c|}{6.55\%}       & \textbf{98.29}                     \\ \hline
Avg. Rank                 & 4.62     & \underline{2.62}    & \multicolumn{1}{c|}{---}                               & 3.85     & 2.85    & \multicolumn{1}{c|}{---}                                 & \textbf{1.08}                       \\ \hline
\end{tabular}
}
}
\end{table}

\subsection{Results of Training Time and Memory Usage Comparisons}\label{section-appendix-comparison-training-time-memory-usage}
\tabref{tab:comparison_training_time_memory_usage_with_without_patching} shows the comparisons of running time and memory usage of DyGFormer with and without the patching technique during the training process.
\begin{table}[!htbp]
\centering
\caption{Comparisons of training time and memory usage of DyGFormer with and without the patching technique, where OOM stands for Out-Of-Memory.}
\label{tab:comparison_training_time_memory_usage_with_without_patching}
\begin{tabular}{c|c|c|cc|c}
\hline
Datasets                    & \begin{tabular}[c]{@{}c@{}}Input \\ Lengths\end{tabular} & Metrics      & \begin{tabular}[c]{@{}c@{}}DyGFormer \\ w/ patching\end{tabular} & \begin{tabular}[c]{@{}c@{}}DyGFormer \\ w/o patching\end{tabular} & \begin{tabular}[c]{@{}c@{}}Reduced \\ Ratios\end{tabular} \\ \hline
\multirow{8}{*}{LastFM}     & \multirow{2}{*}{64}                                      & running time & 13min 58s                                                        & 21min 01s                                                         & 1.50                                                      \\
                            &                                                          & memory usage & 3,945 MB                                                         & 7,953 MB                                                          & 2.02                                                      \\ \cline{2-6} 
                            & \multirow{2}{*}{128}                                     & running time & 16min 54s                                                        & 45min 29s                                                         & 2.69                                                      \\
                            &                                                          & memory usage & 4,677 MB                                                         & 7,585 MB                                                          & 1.62                                                      \\ \cline{2-6} 
                            & \multirow{2}{*}{256}                                     & running time & 24min 41s                                                        & 2h 5min 50s                                                       & 5.10                                                      \\
                            &                                                          & memory usage & 4,635 MB                                                         & 14,583 MB                                                         & 3.15                                                      \\ \cline{2-6} 
                            & \multirow{2}{*}{512}                                     & running time & 37min 04s                                                        & ---                                                               & ---                                                       \\
                            &                                                          & memory usage & 7,547 MB                                                         & OOM                                                               & ---                                                       \\ \hline
\multirow{8}{*}{Can. Parl.} & \multirow{2}{*}{64}                                      & running time & 58s                                                              & 1min 14s                                                          & 1.28                                                      \\
                            &                                                          & memory usage & 2,121 MB                                                         & 3,263 MB                                                          & 1.54                                                      \\ \cline{2-6} 
                            & \multirow{2}{*}{128}                                     & running time & 1min 02s                                                         & 2min 39s                                                          & 2.56                                                      \\
                            &                                                          & memory usage & 2,369 MB                                                         & 6,417 MB                                                          & 2.71                                                      \\ \cline{2-6} 
                            & \multirow{2}{*}{256}                                     & running time & 1min 26s                                                         & 6min 37s                                                          & 4.62                                                      \\
                            &                                                          & memory usage & 2,855 MB                                                         & 14,923 MB                                                         & 5.23                                                      \\ \cline{2-6} 
                            & \multirow{2}{*}{512}                                     & running time & 1min 57s                                                         & ---                                                               & ---                                                       \\
                            &                                                          & memory usage & 4,511 MB                                                         & OOM                                                               & ---                                                       \\ \hline
\end{tabular}
\end{table}

\subsection{Results and Discussions of Ablation Study}\label{section-appendix-ablation-study}
\begin{figure}[!htbp]
    \centering
    \includegraphics[width=1.0\columnwidth]{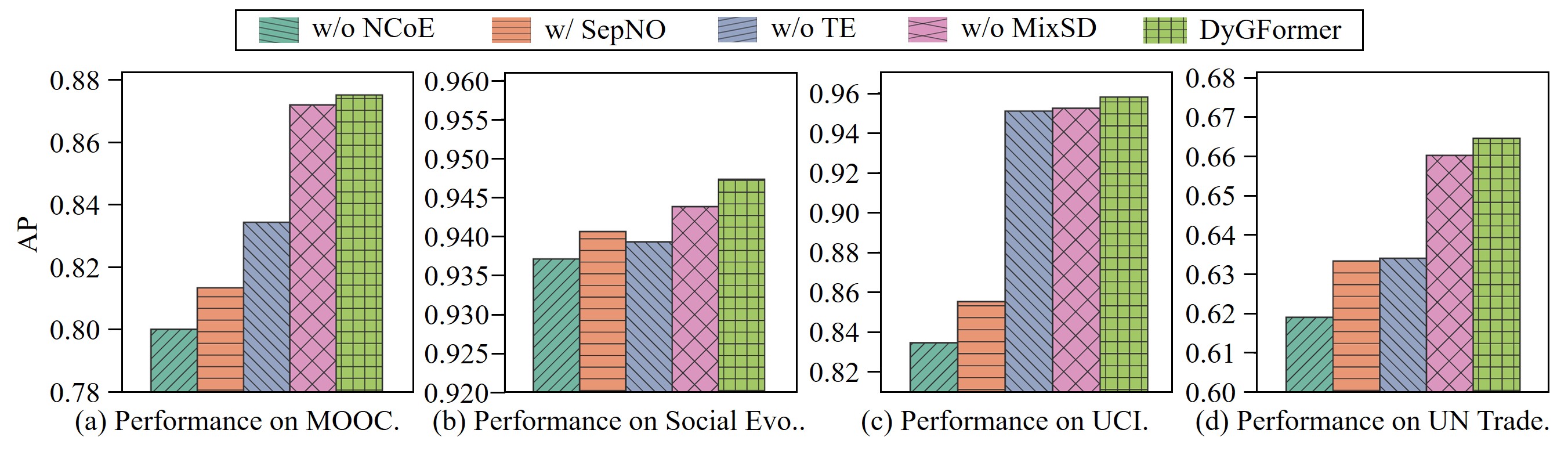}
    \caption{Ablation study of the components in DyGFormer.}
    \label{fig:ablation_study}
\end{figure}

We validate the effectiveness of several designs in DyGFormer via an ablation study, including the usage of \textbf{N}eighbor \textbf{Co}-occurrence \textbf{E}ncoding (NCoE), the usage of \textbf{T}ime \textbf{E}ncoding (TE), and the \textbf{Mix}ing of the sequence of \textbf{S}ource node and \textbf{D}estination node (MixSD). We respectively remove these modules and denote the remaining parts as w/o NCoE, w/o TE, and w/o MixSD. We also \textbf{Sep}arately encode the \textbf{N}eighbor \textbf{O}ccurrence in the source node's or destination node's sequence and denote this variant as w/ SepNO. We report the performance of different variants on MOOC, Social Evo., UCI, and UN Trade datasets from four domains in \figref{fig:ablation_study}.

Our findings from \figref{fig:ablation_study} indicate that DyGFormer achieves optimal performance when all components are utilized. The removal of any component would lead to worse results. In particular, the neighbor co-occurrence encoding scheme has the most significant impact on the performance as it effectively captures correlations between nodes. Separately encoding neighbor occurrences or encoding the temporal information could also improve performance. Mixing the sequences of the source node and destination node causes relatively minor improvements due to the usage of the neighbor co-occurrence encoding scheme because both of them aim to explore the node correlations. To this end, the necessity of designing these components has been well demonstrated.

\subsection{Results on Temporal Graph Benchmark}
We also evaluate DyGFormer and baselines on the recently proposed Temporal Graph Benchmark (TGB) \cite{huang2023temporal}, which contains a collection of challenging and diverse benchmark datasets. We aim to verify the superiority and scalability of DyGFormer on TGB since it covers small, medium, and large datasets. We only present the analysis here. Please refer to \cite{huang2023temporal} and our other work \cite{yu2023empirical} for details of the datasets, tasks, evaluation metrics, experimental settings, and complete quantitative results.

\textbf{Superiority of DyGFormer}. For the dynamic link property prediction task, DyGFormer ranks first on the TGB leaderboard\footnote{\url{https://tgb.complexdatalab.com/docs/leader_linkprop/}} on tgbl-wiki-v2. It ranks second/third on tgbl-coin-v2/tgbl-comment and does medium on tgbl-review-v2. We notice that the surprise index (defined as the ratio of test links that are not seen during training) on tgbl-wiki-v2 and tgbl-coin-v2 are 0.108 and 0.120, which indicates nodes tend to interact repeatedly. Such a property may be well handled by the neighbor co-occurrence encoding scheme in DyGFormer, leading to excellent performance. Correspondingly, the surprise index on tgbl-review-v2 and tgbl-comment are 0.987 and 0.823, showing that these two datasets contain many new links. This characteristic may violate the motivation of our neighbor co-occurrence encoding scheme and bring suboptimal performance. For the dynamic node property prediction task, DyGFormer also performs better than other methods with trainable parameters, but its performance is still worse than the simple non-trainable Persistent Forecast and Moving Average methods on the TGB leaderboard\footnote{\url{https://tgb.complexdatalab.com/docs/leader_nodeprop/}}. This observation demonstrates the necessity of proposing customized models for the dynamic node property prediction task.

\textbf{Scalability of DyGFormer}. DyGFormer can be scaled up to larger datasets that contain hundreds of thousands of nodes and tens of millions of links (e.g., tgbl-coin-v2, tgbl-comment, tgbn-genre, and tgbn-reddit) while many baselines (e.g., JODIE, DyRep, TGN, and CAWN) fail. This demonstrates the good scalability of DyGFormer.

\end{document}